\documentclass{article} % For LaTeX2e
\usepackage[preprint]{neurips_2026}

\usepackage{amsmath,amsfonts,bm}

\def\eqref#1{equation~\ref{#1}}
\def\1{\bm{1}}

\DeclareMathAlphabet{\mathsfit}{\encodingdefault}{\sfdefault}{m}{sl}
\SetMathAlphabet{\mathsfit}{bold}{\encodingdefault}{\sfdefault}{bx}{n}

\newcommand{\E}{\mathbb{E}}

\newcommand{\KL}{D_{\mathrm{KL}}}

\usepackage{hyperref}
\usepackage{url}
\usepackage{graphicx}  
\usepackage{subcaption}
\usepackage{cleveref}
\usepackage{siunitx}
\usepackage{tcolorbox}
\usepackage{wrapfig}
\usepackage{amsmath}
\usepackage{algorithm}
\usepackage{algpseudocode}
\usepackage{booktabs}

\definecolor{takeawaysColor}{HTML}{009E73} 

\title{Replay on Demand: An Emergent Curriculum for Balancing Adaptation and Forgetting in Continued Pretraining}

\author{%
  Lukas Thede$^{1,2,3,4}$ \And
  Shengzhuang Chen$^{4,5}$ \And
  Stefan Winzeck$^{4}$ \AND
  Matthias Bethge$^{1}$ \And
  Zeynep Akata$^{2,3,6}$ \And
  Jonathan Richard Schwarz$^{5}$ \\[1ex]
  $^{1}$University of T\"ubingen, T\"ubingen AI Center \quad
  $^{2}$Helmholtz Munich \\
  $^{3}$Munich Center for Machine Learning (MCML) \quad
  $^{4}$Thomson Reuters Foundational Research \\
  $^{5}$Imperial College London \quad
  $^{6}$Technical University of Munich
}

\begin{document}

\maketitle

\begin{abstract}
    Continued pretraining enables language models to adapt to new domains and knowledge, but often at the cost of forgetting previously acquired capabilities. Replay can mitigate this trade-off, but fixed replay mixtures allocate training independently of the model's actual retention needs. 
    %This mismatch matters because forgetting is heterogeneous across capabilities and data sources, and retention needs evolve during training.
    This is particularly limiting because forgetting varies across capabilities and data sources and evolves throughout training.
    We introduce \emph{Replay on Demand} (RoD), which instead derives the replay allocation from the model's learning dynamics.
    RoD jointly prioritizes adaptation samples by their remaining learning potential and replay samples by their observed forgetting.
    Their competition for a shared training budget yields an online curriculum that determines what to train on at each step.
    Across models, scales, and adaptation domains, RoD reaches or improves upon the adaptation--forgetting frontier of tuned fixed-replay baselines and model merging without prescribing a replay allocation in advance. 
    Replay concentrates on sources that are more vulnerable to forgetting and dynamically increases and redistributes as forgetting emerges during training. Together, our results show that replay can be allocated online from the model's evolving state, targeting \emph{what is needed, when it is needed}.
\end{abstract}

% \begin{figure}[h]
% \centering
% \begin{subfigure}[b]{0.68\textwidth}
% \includegraphics[width=\linewidth]{images/Method_Figure_v2_5.pdf}
% \caption{\textbf{Joint adaptation and replay selection.}
% Adaptation candidates are scored by their remaining learning potential and replay candidates by their forgetting. Joint top-$k$ selection lets both compete for a shared training budget, determining the replay share and composition.}
% \label{fig:method-schematic}
% \end{subfigure}\hfill
% \begin{subfigure}[b]{0.31\textwidth}
% \includegraphics[width=\linewidth]{images/fig1b_mechanism_narrow.pdf}
% \caption{\textbf{Emergent replay allocation.}
% Replay increases as forgetting emerges and adaptation learning potential decreases.}
% \label{fig:method-measured}
% \end{subfigure}
% \caption{\textbf{RoD dynamically balances adaptation and replay through joint data selection.}
% Source-specific reducible-loss signals let the allocation between adaptation and replay emerge from the model's evolving learning and forgetting state.}
% \label{fig:method}
% \vspace{-10pt}
% \end{figure}

\begin{figure}[h]
\centering
\begin{subfigure}[b]{0.68\textwidth}
\includegraphics[width=\linewidth]{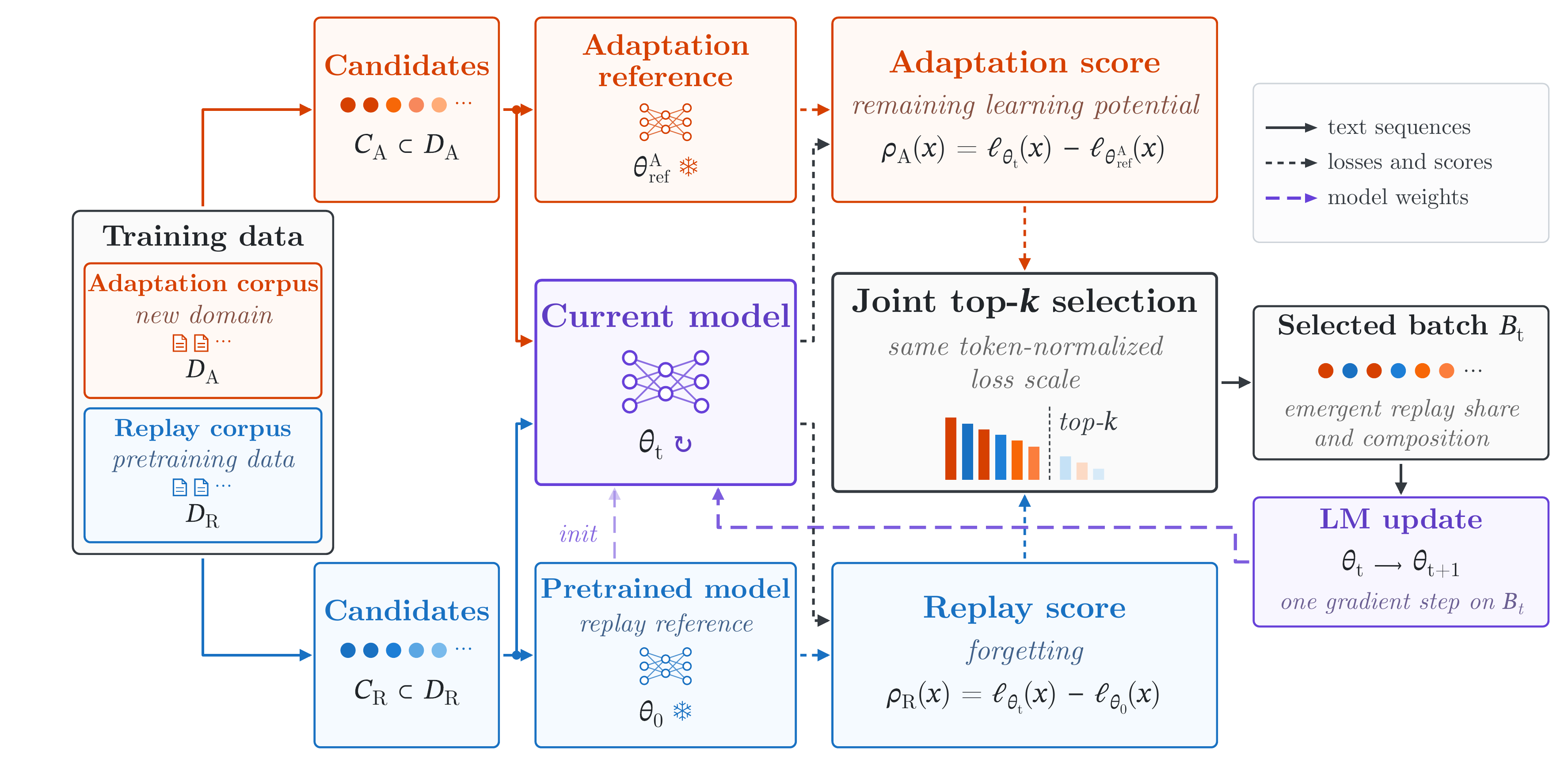}
\label{fig:method-schematic}
\end{subfigure}\hfill
\begin{subfigure}[b]{0.31\textwidth}
\includegraphics[width=\linewidth]{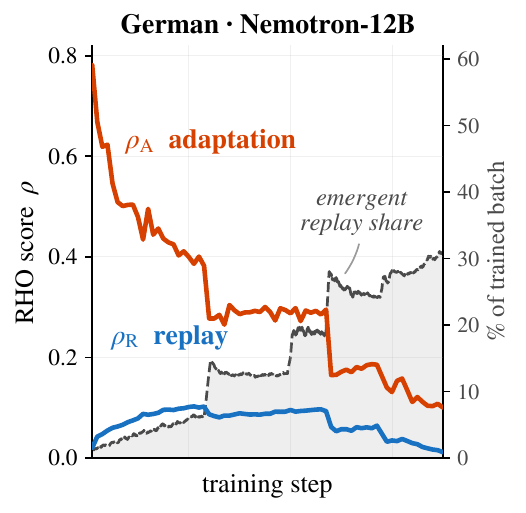}
\label{fig:method-measured}
\end{subfigure}
\vspace{-10pt}
\caption{\textbf{RoD dynamically balances adaptation and replay through joint data selection.}
Adaptation candidates are scored by their remaining learning potential and replay candidates by their forgetting. Joint top-$k$ selection lets both compete for a shared training budget, dynamically determining the replay share and composition. As adaptation learning potential decreases and forgetting emerges, replay becomes increasingly competitive and receives a larger share of the training budget.}
\label{fig:method}
\vspace{-10pt}
\end{figure}

\section{Introduction}
\label{sec:introduction}

Continued pretraining (CPT) has become a practical route for adapting general-purpose language models to specialized domains, languages, and applications~\citep{gururangan-etal-2020-dont,parmar2024reusedontretrainrecipe}.
However, this specialization comes at a cost: learning a new distribution can degrade capabilities acquired during pretraining.
CPT therefore faces an inherent trade-off between \emph{adaptation} and \emph{retention}: a successful method should adapt to the new distribution while minimizing forgetting of the original pretraining distribution.

A common strategy for controlling this trade-off is \emph{replay}, which mixes samples from the original pretraining distribution into the adaptation data~\citep{pmlr-v330-abbes26a,roth2024practitionersguidecontinualmultimodal}.
Yet replay turns this trade-off into a resource-allocation problem.
Under a fixed training budget, replay displaces adaptation data, while maintaining both requires additional compute.
Replay should therefore be allocated where it provides the greatest retention benefit.
This is challenging because forgetting is heterogeneous.
Different capabilities and parts of the pretraining distribution degrade to different degrees during adaptation~\citep{thede2026captrackmultifacetedevaluationforgetting,yildiz2024investigating}.
Fixed replay mixtures ignore this heterogeneity, potentially spending compute on well-retained knowledge while leaving more vulnerable knowledge insufficiently protected.
Moreover, retention needs evolve throughout training.
An effective replay curriculum should therefore allocate replay where forgetting occurs and adapt this allocation as retention needs change.

% A common strategy for controlling this trade-off is \emph{replay}: mixing samples from the original pretraining distribution into the adaptation data~\citep{abbes2026revisiting,roth2024practitionersguidecontinualmultimodal}. Yet replay turns the adaptation–retention trade-off into a resource-allocation problem because replay is not free. Under a fixed training budget, every replayed example displaces adaptation data; maintaining both instead requires additional training compute. Replay should therefore be allocated where it provides the greatest retention benefit. This is particularly important because forgetting in language models is heterogeneous, with different capabilities and parts of the pretraining distribution degrading to different degrees during adaptation~\citep{thede2026captrackmultifacetedevaluationforgetting, yildiz2024investigating}. Fixed replay mixtures ignore this heterogeneity by allocating retention compute independently of what the model actually forgets. As a result, replay may be spent on well-retained knowledge while more vulnerable knowledge remains insufficiently protected. Moreover, these retention needs evolve as adaptation progresses. 
% An effective replay curriculum should therefore allocate replay according to \emph{where} forgetting occurs and continuously adjust this allocation as retention demands change. 

We propose \emph{Replay on Demand} (RoD), which allocates training compute based on the model's current learning and forgetting states. RoD builds on reducible-loss data selection~\citep{mindermann2022prioritized} and lets adaptation and replay examples compete for a shared training budget. Adaptation examples are prioritized by how much remains to be learned, while replay examples are prioritized by how much has been forgotten relative to the pretrained model. This competition gives rise to an online data curriculum. When pretrained knowledge is well retained, adaptation dominates training. As forgetting emerges, affected replay examples become more competitive and enter the training batch. RoD thereby concentrates replay on the parts of the pretraining distribution that degrade, dynamically adjusting its amount and composition throughout training. The resulting curriculum adapts to the model and adaptation corpus without requiring a predefined replay share or allocation.

% We propose \emph{Replay on Demand} (RoD), which allocates training compute according to the model’s current adaptation and retention needs. At each training step, RoD draws candidate examples from both the adaptation and replay distributions and lets them compete for a shared training budget. We build on reducible-loss data selection~\citep{mindermann2022prioritized}, using source-specific reference models to quantify the demand for each example. For adaptation data, reducible loss measures how much remains to be learned. For replay data, we instead use the original pretrained model as the reference, such that the loss difference measures how much has been forgotten. 
% Jointly ranking both types of candidates therefore allocates compute between adaptation and retention based on the model’s current state, rather than on a predefined replay mixture. 

% This competition gives rise to a demand-driven replay curriculum. When pretrained knowledge is well retained, adaptation examples dominate the training budget. As forgetting emerges, affected replay examples become more competitive and receive more compute. Consequently, RoD determines not only \emph{how much} to replay, but also \emph{what} to replay and \emph{when}: replay is concentrated on the parts of the pretraining distribution that degrade, at the stages of training when they require protection. The resulting curriculum adapts to the model and adaptation corpus without requiring a predefined replay share or allocation.

We evaluate RoD across legal and German continued pretraining using the Nemotron and Qwen model families at scales ranging from 4B to 35B parameters. Across these settings, RoD reaches or improves upon the adaptation–forgetting frontier of tuned fixed-replay CPT and model merging, without prescribing a replay allocation in advance. Our analysis shows that these gains reflect the intended demand-driven allocation: RoD directs replay toward sources that exhibit greater forgetting, avoids unnecessary replay of well-retained ones, and dynamically increases and redistributes replay as retention needs emerge during training. 
Together, these results show that adapting replay to the model’s evolving retention needs provides a principled way to balance adaptation and retention by replaying what is needed, when it is needed.

% Our contributions are:
% \begin{itemize}
%     \item We introduce \textbf{Replay on Demand (RoD)}, a continual-pretraining method that jointly selects adaptation and replay data based on the model's current adaptation and retention needs, yielding a demand-driven \textbf{data curriculum} without a predefined replay allocation.
    
%     \item We show that RoD reaches or improves the \textbf{adaptation--forgetting frontier} of tuned fixed-replay CPT and model merging at matched trained-token budgets across domains, model families, and scales up to 35B parameters.
    
%     \item We show that RoD allocates replay according to \textbf{evolving retention needs}: replay is directed toward sources that forget more, while its amount and composition dynamically adjust as forgetting emerges during training.
% \end{itemize}

\section{Related Work}
\label{sec:related_work}

\paragraph{Continual pretraining and replay.}
CPT adapts pretrained language models to new data while aiming to preserve previously acquired capabilities~\citep{gururangan-etal-2020-dont,yildiz2024investigating}.
Replay of pretraining data is a common strategy for mitigating forgetting~\citep{ibrahim2024simple,parmar2024reusedontretrainrecipe,roth2024practitionersguidecontinualmultimodal}, making the allocation of training compute between adaptation and retention a central design choice.
Recent work has studied this trade-off through scaling laws for source–target mixture ratios~\citep{gu2024cmr,que2024d}, adaptive data mixtures~\citep{chen-etal-2025-towards-effective,luo-etal-2025-velocitune,yang2026data}, and replay mechanisms that select or schedule past data based on their utility or the model’s learning state~\citep{atreya2026spaced,feng2026forever,smith2024adaptive}.
Correspondingly, CPT evaluation considers both adaptation and retention rather than target-domain performance alone~\citep{jin-etal-2022-lifelong-pretraining,li-etal-2025-tic}.
We build on this perspective but ask whether this allocation can emerge from the model’s learning and forgetting.

% \paragraph{Reducible-loss data selection.}
% Reducible Holdout Loss (RHO) was introduced for data-efficient training, prioritizing examples according to how much of their current loss remains reducible relative to a reference model~\citep{mindermann2022prioritized}.
% Sequence RHO extends this principle to sequence-level selection for language-model pretraining~\citep{thirukovalluru2024sequence}, while Rho-1 applies a similar
% reference-model-guided approach at the token level, selectively computing loss only on informative
% tokens during pretraining~\citep{lin2024rho}.
% More broadly, model-dependent signals have been used to dynamically adapt pretraining mixtures~\citep{tikmix}, while CSReL applies reducible-loss selection specifically to replay examples in continual learning~\citep{Tong2025CoresetSV}.
% These methods use the model’s state to determine which data to train on, but do not jointly allocate compute between learning new data and retaining previously learned knowledge.

\paragraph{Reducible-loss data selection.}
Reducible Holdout Loss (RHO) was introduced for data-efficient training, prioritizing examples according to how much of their current loss remains reducible relative to a reference model~\citep{mindermann2022prioritized}.
Subsequent work extends reference-model-guided selection to sequences~\citep{thirukovalluru2024sequence} and tokens~\citep{lin2024rho} in language-model pretraining.
Related work uses model-dependent signals to adapt pretraining mixtures~\citep{tikmix}, while CSReL applies reducible-loss selection to replay examples in continual learning~\citep{Tong2025CoresetSV}.
These methods determine which data to train on based on the model’s state, but do not jointly allocate compute between adaptation and retention.

\paragraph{Positioning RoD.}
RoD connects adaptive CPT and model-dependent data selection by placing adaptation and replay examples in a shared competition. 
Unlike methods optimizing a source–target mixture~\citep{chen-etal-2025-towards-effective,gu2024cmr,luo-etal-2025-velocitune,que2024d,yang2026data} or select replay within a dedicated mechanism~\citep{atreya2026spaced,feng2026forever,smith2024adaptive,Tong2025CoresetSV}, RoD jointly determines replay amount and composition from the relative utility of adaptation and retention candidates. Its reference models encode these objectives: the adaptation reference estimates learnability, while the pretrained model measures degradation. The adaptation–retention trade-off thus emerges from the model’s evolving state rather than a predefined mixture or separate replay mechanism. Controlled fixed-allocation baselines isolate this distinction by prescribing the replay allocation while retaining model-dependent selection (\cref{sec:ablations}).

\section{Method}
\label{sec:method}

We introduce \emph{Replay on Demand} (RoD), an online data-selection method that dynamically allocates training compute between adaptation and replay.
As illustrated in \cref{fig:method}, RoD scores adaptation and replay examples with source-specific reducible losses and lets them compete for a shared training budget.
The resulting selection determines both \emph{how much} and \emph{what} to replay throughout training.

\subsection{Problem Setup}
\label{sec:method_setup}

Let $\theta_0$ denote a pretrained language model that we adapt to a new data distribution $\mathcal{D}_{A}$ while retaining performance on its pretraining distribution $\mathcal{D}_{R}$.
During continual pretraining, $\theta_t$ denotes the model after $t$ updates.
We construct candidate sets
\begin{equation}
C_A \subset \mathcal{D}_{A},
\qquad
C_R \subset \mathcal{D}_{R},
\end{equation}
and select a training batch $B_t$ of $k$ examples from their union.
We denote the token-normalized language-modeling loss of model $\theta$ on an example $x$ by $\ell_\theta(x)$.

Rather than fixing the fraction of adaptation and replay data in $B_t$ beforehand, RoD lets candidates from both sources compete for the same training budget based on their current utility to the model.

\subsection{Source-Specific Reducible Loss}
\label{sec:method_scores}

RoD builds on Reducible Holdout Loss (RHO) \citep{mindermann2022prioritized}, which scores examples by the difference between their current loss and the loss of a reference model:
\begin{equation}
\rho(x;\theta_t)
=
\ell_{\theta_t}(x)
-
\ell_{\theta_{\mathrm{ref}}}(x).
\label{eq:rho}
\end{equation}
The reference loss estimates how well an example can be explained, such that the difference captures the loss that remains reducible by further learning.
RHO therefore prioritizes examples that the current model does not yet explain well, but that a suitable reference model does.

In continual pretraining, adaptation should target what remains to be learned, while replay targets what has been forgotten. We capture both with the same reducible-loss formulation, choosing the reference model according to the candidate source:
\begin{equation}
\rho(x;\theta_t)
=
\begin{cases}
    \rho_A(x;\theta_t)
    = \ell_{\theta_t}(x) - \ell_{\theta_{\mathrm{ref}}^A}(x),
    & x \in C_A, \\
    \rho_R(x;\theta_t)
    = \ell_{\theta_t}(x) - \ell_{\theta_0}(x),
    & x \in C_R.
\end{cases}
\label{eq:RoD}
\end{equation}
where $\theta_{\mathrm{ref}}^A$ is the adaptation reference model specialized to the adaptation distribution and the pretrained model $\theta_0$ serves as the replay reference.
Both scores therefore measure excess loss relative to a source-appropriate reference, providing a common scale for joint selection.

This construction follows from a source-specific approximation of the original RHO objective, which measures how much training on an example improves the posterior predictive likelihood of a holdout distribution. In CPT, the holdout contains both adaptation and retained pretraining data. Approximating its posterior predictive separately for each source yields $\theta_{\mathrm{ref}}^A$ for adaptation and $\theta_0$ for replay. A single reference trained on both would instead encode an adaptation–retention trade-off through its training mixture. We provide the full derivation and assumptions in \cref{app:rho_derivation}.

\paragraph{Adaptation: remaining learning potential.}
For adaptation examples, $\theta_{\mathrm{ref}}^A$ estimates the loss attainable after learning the adaptation distribution, without requiring the reference to retain the general capabilities of $\theta_0$.
The resulting $\rho_A$ is high for examples that the current model does not yet explain well but the adaptation specialist does.
As the model learns these examples, their scores decrease.
Thus, $\rho_A$ prioritizes remaining learning potential on $\mathcal{D}_A$.

\paragraph{Replay: forgetting from the pretrained state.}
For replay examples, the pretrained model $\theta_0$ provides a natural reference for detecting degradation.
Here, $\rho_R$ directly measures the change in loss since the start of adaptation.
Retained examples have $\rho_R(x;\theta_t)\approx 0$, whereas examples whose performance has degraded obtain positive scores.
Replay therefore prioritizes knowledge in proportion to its currently observed forgetting.

\subsection{Joint Adaptation and Replay Selection}
\label{sec:method_selection}

Given candidate sets $C_A$ and $C_R$, RoD scores each candidate according to its source and selects the globally highest-scoring $k$ examples:
\begin{equation}
B_t
=
\operatorname{TopK}_{x\in C_A\cup C_R}
\rho(x;\theta_t).
\label{eq:joint_selection}
\end{equation}
The model is then updated on $B_t$ using the standard language-modeling objective. Under a local marginal-utility approximation, this selection allocates each training slot to candidates with the largest expected reduction in their source objective. The replay share thus emerges from the relative demand of adaptation and replay rather than a predefined ratio. We formalize this interpretation and its assumptions in \cref{app:joint_selection_theory}.

% We control the size of the candidate pool via a candidate multiplier $m$, with $|C_A|+|C_R|=mk$.
% In our implementation, $m$ is the primary selection hyperparameter and determines how many candidates compete for each training slot.
% Larger values of $m$ provide greater flexibility in selecting both the composition and individual examples of the training batch, at the cost of scoring more candidates.
% We use $m=2$ by default and study its effect in Section~\ref{sec:ablations}.
We control the size of the candidate pool via a candidate multiplier $m$, with $|C_A|+|C_R|=mk$. Larger $m$ allows more candidates to compete for each training slot, providing greater flexibility in selecting the batch at the cost of additional scoring. We use $m=2$ by default and study its effect in Section~\ref{sec:ablations}.

Figure~\ref{fig:method}b illustrates how this competition produces a dynamic replay allocation.
At the start of adaptation, $\theta_t=\theta_0$ and all replay scores are zero, such that adaptation examples dominate selection.
As adaptation progresses, scores decrease for learned adaptation examples, while replay scores increase wherever pretrained knowledge degrades.
Replay consequently becomes more competitive and receives a larger share of the training budget.
Within the replay pool, examples with greater degradation similarly outrank well-retained examples.
RoD thereby adapts both the replay share and replay composition to the model's evolving learning and forgetting state.

\paragraph{Practical implementation.}
Reference-model losses are fixed throughout continual pretraining and can therefore be precomputed and cached for both candidate sources.
Each training step then requires one inference pass of the current model over the candidate pool for scoring, followed by a standard training pass over the selected batch.
While RoD does not require equal candidate-pool sizes, we balance adaptation and replay candidates in our experiments so that $m=2$ allows the selected batch to consist entirely of either source.
We obtain $\theta_{\mathrm{ref}}^A$ by training the pretrained model on $\mathcal{D}_A$ until convergence, without replay or other forgetting mitigation.
Further algorithmic and implementation details are provided in \cref{app:algorithm,app:adaptation_reference}.
Experimental details are provided in \cref{app:experimental_setup}, and computational requirements and limitations are discussed in \cref{app:limitations}.
Our implementation builds on NeMo-RL \citep{nemo-rl}, with code for RoD and our experimental setup available at \url{https://github.com/bethgelab/replay-on-demand}.

\section{Experiments}
\label{sec:experiments}

We evaluate RoD across two model families and two adaptation domains, progressively testing the generality of our approach.
We evaluate Nemotron-Nano-12B-v2~\citep{nvidia2025nvidianemotronnano2} on Legal and German with access to its original pretraining data for replay, and Qwen3.5-9B~\citep{qwen3.5} on both domains to test generalization across model families and to proxy replay.
For our cross-scale experiments in \cref{sec:cross_scale}, we additionally use Qwen3.5-4B as a smaller source model and transfer curricula constructed at smaller scale to Nemotron-30B~\citep{nvidia2026nemotron3ultraopen} and Qwen3.5-35B as larger target models.
Full experimental details are provided in \cref{app:experimental_setup}.

We compare RoD against the no-replay baseline, CPT with several fixed replay shares, and post-hoc model merging between the base and no-replay CPT models.
%RoD and fixed-replay CPT use matched trained-token budgets and the same learning-rate schedule.
RoD and fixed-replay CPT use the same learning-rate schedule and matched trained-token budgets, while RoD incurs additional candidate-scoring overhead (see~\cref{app:computational_requirements}).
We additionally train a max-replay reference on the full available adaptation–replay mixture using approximately $2\times$ the trained-token budget, and therefore treat it as a higher-budget reference rather than part of the matched-budget comparison.

We measure adaptation by validation loss on held-out adaptation data and forgetting by the increase in held-out general validation loss relative to the base model.
We complement this distribution-level evaluation with a capability-level perspective, using multiple-choice evaluations of four latent capability groups following the CapTrack taxonomy~\citep{thede2026captrackmultifacetedevaluationforgetting}: parametric knowledge, reasoning and problem-solving, commonsense and robustness, and multilingual capabilities.
Further details on data construction, training, baselines, and evaluation are provided in \cref{app:experimental_setup}.

\subsection{RoD improves the adaptation--forgetting frontier}
\label{sec:adaptation_forgetting_frontier}

\begin{figure}[t]
  \centering
  \includegraphics[width=\textwidth]{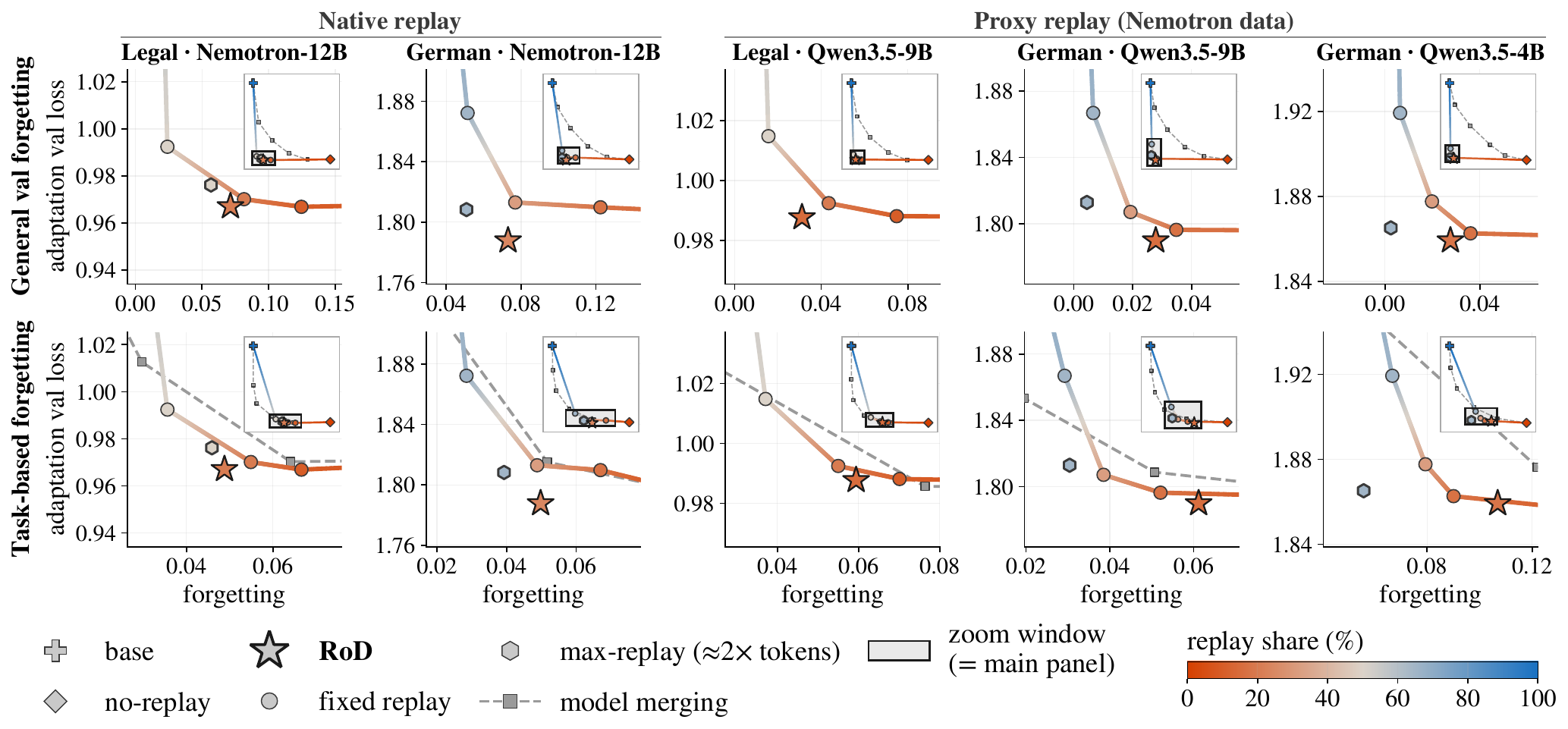}
  \caption{\textbf{RoD reaches or improves upon the adaptation--forgetting frontier.}
    Adaptation loss is shown against general validation-loss forgetting (top) and task-based forgetting (bottom); lower is better on both axes. Fixed-replay CPT traces the frontier as replay share varies, while RoD reaches or improves upon it without specifying a replay ratio in advance. Nemotron uses native replay from its pretraining data, whereas Qwen uses the Nemotron data as proxy replay because its pretraining data are unavailable. No-replay CPT and model merging provide additional baselines; max replay is a higher-budget reference trained with approximately $2\times$ the trained-token budget. Insets magnify the operating region around RoD and the strongest baselines.}
  \label{fig:frontier}
\vspace{-10pt}
\end{figure}

%We characterize the stability–plasticity trade-off through the adaptation–forgetting frontier in \cref{fig:frontier}, where desirable methods combine strong adaptation with low validation-loss and task-based forgetting. Fixed-replay CPT traces this frontier as the replay share varies: more replay reduces forgetting but limits target-domain adaptation. Since the preferred balance is setting-dependent, selecting a fixed replay ratio requires prior knowledge or a sweep over ratios. We therefore use the full sweep as a reference frontier, noting that selecting its best point for comparison constitutes an oracle-like choice. Complete numerical results and fixed-replay baselines trained to convergence are provided in \cref{app:extended_frontier}.
We characterize the stability–plasticity trade-off through the adaptation–forgetting frontier in \cref{fig:frontier}, where desirable methods combine strong adaptation with low validation-loss and task-based forgetting. Fixed-replay CPT traces this frontier as replay varies: more replay reduces forgetting but limits target-domain adaptation. Since the preferred balance is setting-dependent, selecting a fixed replay ratio requires prior knowledge or a sweep over ratios. We use the full sweep as a reference frontier, noting that selecting its best point for comparison constitutes an oracle-like choice. Complete numerical results and fixed-replay baselines trained to convergence are provided in \cref{app:extended_frontier}.

\paragraph{Does RoD improve the frontier with native replay?}
Across both Nemotron settings, RoD reaches or improves upon the fixed-replay frontier without specifying a replay ratio in advance.
At matched adaptation on Legal, RoD reduces validation-loss forgetting by 43\% relative to 10\% fixed replay (0.125 to 0.071) and task-based forgetting from 6.7 to 4.9 points.
On German, RoD maintains substantially stronger adaptation while closely matching the retention of the more replay-heavy 32\% baseline.
RoD also compares favorably with alternative strategies: model merging achieves stronger adaptation only at substantially higher forgetting, whereas max-replay training yields slightly better retention with approximately twice the trained-token budget.
Despite this additional budget, RoD achieves better adaptation than max replay in both settings while remaining close in retention.
Thus, a single RoD run achieves a strong adaptation--forgetting trade-off without tuning a replay ratio or training exhaustively on the available replay data.

\paragraph{Does RoD remain effective with proxy replay?}
For Qwen3.5, whose original pretraining data are unavailable, we instead use the Nemotron data as proxy replay.
RoD continues to reach or improve upon the validation-loss frontier across both domains and model scales.
On Legal, Qwen3.5-9B RoD matches the adaptation loss of 10\% fixed replay (0.988) while reducing validation-loss forgetting from 0.075 to 0.031.
%On German, it closely matches 17\% fixed replay in adaptation loss (1.790 vs.\ 1.796) while achieving low validation-loss forgetting (0.028 vs.\ 0.035), with the same qualitative pattern for Qwen3.5-4B.
On German, it achieves a comparable trade-off to 17\% fixed replay (1.790 vs.\ 1.796 adaptation loss; 0.028 vs.\ 0.035 forgetting), with the same pattern at 4B. Higher replay shares further reduce forgetting at the cost of adaptation.

For task-based forgetting, RoD's advantage over fixed replay is smaller, which we attribute to the proxy replay loss being less well aligned with retention of Qwen's evaluated capabilities. We analyze this discrepancy in \cref{app:qwen_proxy_replay} and discuss the requirements on the replay distribution and signal in \cref{app:replay_distribution,app:replay_signal}.
Nevertheless, RoD substantially reduces task-based forgetting relative to no-replay CPT: by 60\% on Legal Qwen3.5-9B and by 43\% and 46\% on German Qwen3.5-9B and 4B, respectively, at comparable adaptation.

% The task-based frontier is less pronounced for Qwen, particularly for reasoning, mathematics, and multilingual capabilities.
% We attribute this discrepancy to proxy replay: because the Qwen base models are not loss-converged on the Nemotron replay corpus, base-relative loss on this data is less well aligned with retention of the evaluated capabilities.
% We analyze this effect in \cref{app:qwen_proxy_replay} and discuss the corresponding requirements on the replay distribution and signal in \cref{app:replay_distribution,app:replay_signal}.
% Nevertheless, RoD substantially reduces task-based forgetting relative to no-replay CPT: by 60\% on Legal Qwen3.5-9B and by 43\% and 46\% on German Qwen3.5-9B and 4B, respectively, at comparable adaptation.
% Complete results and comparisons with fixed-replay baselines trained to convergence are provided in \cref{app:extended_frontier}.

\begin{tcolorbox}[
  colback=white,
  colframe=takeawaysColor,
  boxrule=1.5pt,
  arc=3pt,
  left=4pt,right=4pt,top=4pt,bottom=4pt,
]
\textbf{\textcolor{takeawaysColor}{Takeaways.}}
RoD consistently reaches or improves upon the adaptation--forgetting frontier without choosing a replay ratio in advance. A single RoD run thereby reaches trade-offs that fixed-replay CPT obtains by explicitly sweeping the replay allocation.
\end{tcolorbox}

\subsection{RoD learns a demand-driven replay curriculum}
\label{sec:demand_driven_replay}

\begin{figure}[t]
  \centering

  \begin{subfigure}[t]{0.49\textwidth}
    \centering
    \includegraphics[width=\linewidth]{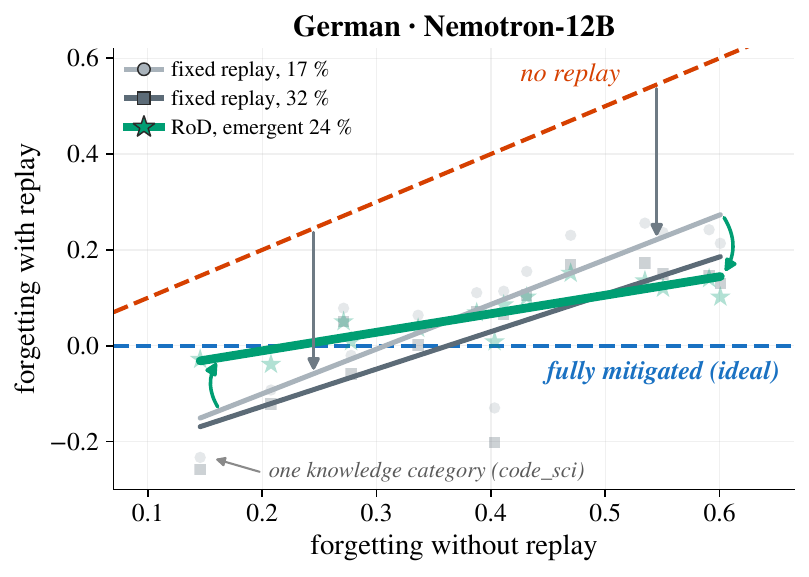}
    \caption{\textbf{Forgetting mitigation.}
    Fixed replay largely preserves the dependence of remaining forgetting on knowledge category vulnerability, whereas RoD substantially flattens this relationship and provides stronger protection where forgetting is greatest.}
    \label{fig:forgetting_profile}
  \end{subfigure}
  \hfill
  \begin{subfigure}[t]{0.49\textwidth}
    \centering
    \includegraphics[width=\linewidth]{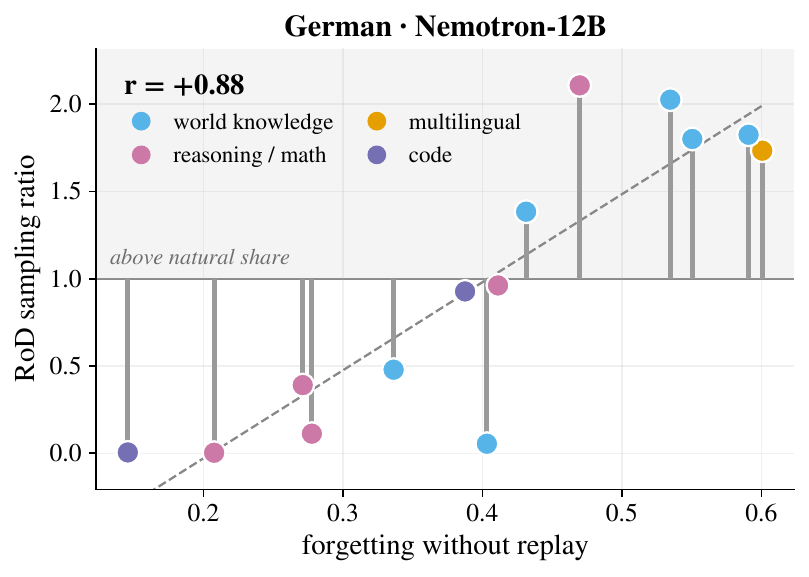}
    \caption{\textbf{Replay allocation.}
    Knowledge categories with greater forgetting under no-replay CPT are sampled more frequently relative to their prevalence in the replay pool, while well-retained categories are sampled less frequently ($r=0.88$).}
    \label{fig:oversampling}
  \end{subfigure}

  \caption{\textbf{RoD allocates replay according to forgetting demand.}
  On German adaptation with Nemotron-12B, RoD concentrates protection on the knowledge categories most vulnerable to forgetting (a) by allocating more replay to these categories during training (b).}
  \label{fig:demand_driven_replay}
\vspace{-10pt}
\end{figure}

The previous results show that RoD reaches a favorable adaptation--forgetting trade-off without specifying a replay ratio. We next examine the resulting curriculum: does RoD replay \emph{what} is needed, \emph{when} it is needed?

\paragraph{Does RoD replay what is being forgotten?}
We first ask whether RoD preferentially protects the parts of the general distribution most vulnerable to forgetting. \Cref{fig:forgetting_profile} relates forgetting under no-replay CPT to the forgetting remaining after replay for each knowledge category in the replay data. Fixed replay largely pushes down the no-replay forgetting profile: increasing the replay share reduces forgetting across categories, but those that forget more without replay continue to forget more after replay. 
In other words, fixed replay spends the same predetermined replay capacity irrespective of source-specific retention need, including on categories that would remain stable with substantially less replay.
RoD instead rotates this profile toward the no-forgetting line, reducing its slope from 0.93 and 0.78 for fixed replay at 17\% and 32\% to 0.39 at an emergent replay share of 24\%. Thus, RoD provides little additional protection where pretrained performance is already retained and concentrates its replay budget where adaptation causes substantial forgetting. We observe the same qualitative behavior across the remaining model--domain settings (\cref{app:demand_driven_extended}).

\Cref{fig:oversampling} shows how this protection emerges from the replay allocation. It relates each category's vulnerability to its sampling ratio under RoD, with values above 1 indicating that a category is sampled more frequently than its prevalence in the replay pool. Categories with greater forgetting are systematically sampled more frequently relative to their share in the replay pool, while well-retained categories are sampled less frequently. For German adaptation of Nemotron-12B, this relationship is strong ($r=0.88$): the most vulnerable web-scale knowledge categories are sampled at up to $2.1\times$ their prevalence in the replay pool, whereas well-retained categories such as math textbooks are sampled less frequently than their share. 
We observe the same positive relationship across all remaining model–domain settings (\cref{app:demand_driven_extended}).
%The same relationship holds across the remaining model--domain settings, with positive correlations of $r=0.76$, $0.66$, and $0.45$ for Legal Nemotron-12B, German Qwen3.5-9B, and German Qwen3.5-4B, respectively (\cref{app:demand_driven_extended}). 
Together, these results show that RoD allocates its replay budget according to what the model is forgetting, rather than simply reproducing the composition of the replay pool.

\begin{figure}[t]
  \centering
  \includegraphics[width=\textwidth]{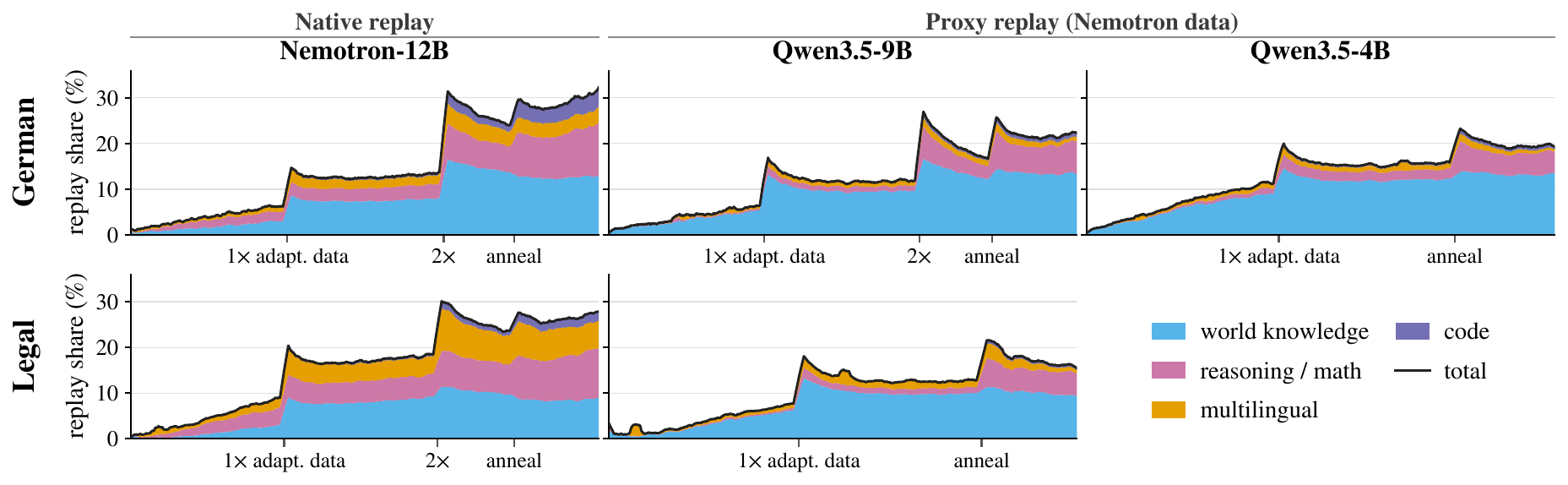}
  \caption{\textbf{RoD dynamically determines what to replay, how much, and when.}
Stacked areas show the fraction of each training batch allocated to different knowledge categories, while the line shows the total replay share. Replay emerges as forgetting develops and is rebalanced throughout training, while its composition evolves across knowledge categories and model--domain settings. RoD thus adapts both the amount and composition of replay over training.}
  \label{fig:curriculum}
\vspace{-10pt}
\end{figure}

\paragraph{When and how much does RoD replay?}
Beyond deciding \emph{what} to replay, RoD continuously adjusts the replay allocation as training progresses. \Cref{fig:curriculum} shows that replay is initially absent because the target and base models coincide, giving replay samples zero reducible loss by construction. Early training therefore focuses predominantly on adaptation, while replay gradually increases as the target distribution is learned and forgetting emerges. This rebalancing is particularly evident around the $1\times$ and $2\times$ markers, which denote cumulative adaptation-data exposure equivalent to one and two dataset sizes.\footnote{Training horizons differ across settings because we match them to the convergence of the corresponding no-replay run. Consequently, not all runs reach $2\times$ adaptation-data exposure.} As training progresses, the remaining learning potential of adaptation samples decreases on average, making replay increasingly competitive under joint selection. RoD consequently shifts toward replay as further adaptation becomes less valuable, reaching replay shares of 19--32\% toward the end of training. 

RoD simultaneously adapts \emph{what} is replayed. The composition within the replay share changes throughout training and differs substantially across model--domain settings. For example, world knowledge accounts for most replay throughout the Qwen runs, whereas reasoning, mathematics, and multilingual data receive substantially larger shares for Nemotron, particularly later in training. These allocations themselves also evolve over time rather than remaining proportional to the replay pool, providing a temporal view of the source-specific replay allocation observed above.

\begin{tcolorbox}[
  colback=white,
  colframe=takeawaysColor,
  boxrule=1.5pt,
  arc=3pt,
  left=4pt,right=4pt,top=4pt,bottom=4pt,
]
\textbf{\textcolor{takeawaysColor}{Takeaways.}}
RoD continuously balances adaptation against retention and retention demands within the replay distribution, determining \emph{what} to replay, \emph{how much}, and \emph{when} based on the model's evolving learning and forgetting dynamics.
\end{tcolorbox}

\subsection{Joint adaptation--replay competition drives RoD}
\label{sec:ablations}

\begin{figure}[t]
\centering

\begin{subfigure}[b]{0.32\textwidth}
  \centering
  \includegraphics[width=\linewidth]{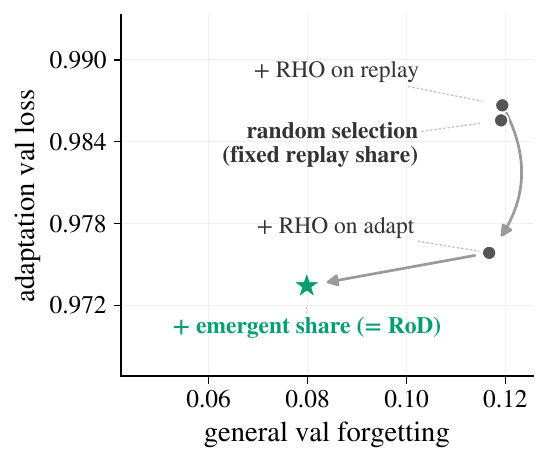}
\caption{\textbf{Component ablation.}
Joint competition improves upon selection at fixed replay allocation.}
  \label{fig:ablation-components}
\end{subfigure}\hfill
\begin{subfigure}[b]{0.32\textwidth}
  \centering
  \includegraphics[width=\linewidth]{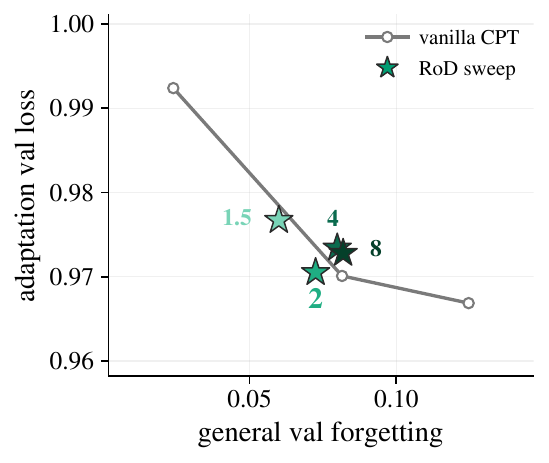}
    \caption{\textbf{Candidate pool size.}
    The trade-off stabilizes once $m\geq2$ leaves replay unconstrained.}
  \label{fig:ablation-multiplier}
\end{subfigure}\hfill
\begin{subfigure}[b]{0.32\textwidth}
  \centering
  \includegraphics[width=\linewidth]{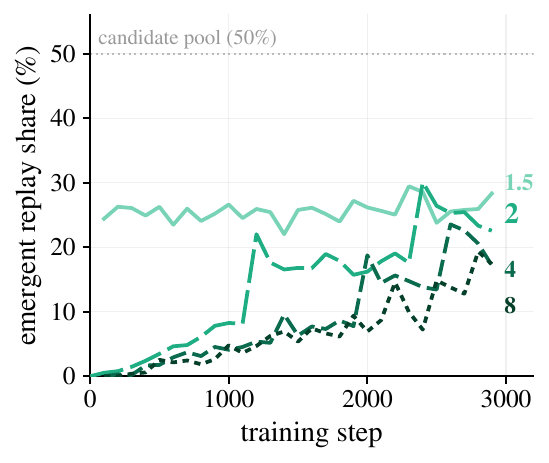}
\caption{\textbf{Emergent replay share.}
Replay trajectories stabilize for $m\geq2$ under unconstrained allocation.}
  \label{fig:ablation-replay-share}
\end{subfigure}

\caption{\textbf{Joint adaptation--replay competition drives RoD's gains.}
Joint competition improves over fixed-allocation selection, while trade-off
and replay allocation stabilize when unconstrained.}
\label{fig:ablation}
\vspace{-10pt}
\end{figure}

Having established how RoD's replay curriculum emerges, we next ask whether its gains arise from RHO-based data selection within a fixed replay allocation or from allowing the allocation itself to emerge. To isolate these effects, \cref{fig:ablation}a constructs a sequence of controlled fixed-allocation baselines that progressively introduce RoD's selection mechanisms. Starting from random selection, we first select replay examples by their RHO score and then apply RHO-based selection to both replay and adaptation examples, while keeping the replay share fixed. These baselines capture increasingly adaptive selection within a predefined replay allocation. RoD additionally removes this constraint and lets adaptation and replay examples compete jointly for the shared training budget.

%Having established how RoD's replay curriculum emerges, we isolate which components drive its adaptation--forgetting trade-off. \Cref{fig:ablation}a progressively introduces RHO-based selection for replay samples, adaptation samples, and finally joint selection with an emergent replay share. Within a fixed replay budget, selecting adaptation samples based on their RHO score improves adaptation, whereas RHO-based selection of replay samples largely maintains forgetting. Allowing adaptation and replay samples to compete jointly further improves adaptation and substantially reduces forgetting, despite the same average replay share. Thus, RoD benefits from both selecting informative adaptation samples and allowing the replay signal to determine which replay samples are sufficiently important to displace adaptation samples from the shared training budget.

Within a fixed replay budget, RHO-based replay selection largely maintains the adaptation--forgetting trade-off of random selection, while selecting adaptation examples improves adaptation. Allowing adaptation and replay examples to compete jointly further improves adaptation and substantially reduces forgetting, despite the same average replay share. Thus, RoD's gains cannot be explained by RHO-based selection within either stream alone. Instead, the gains emerge when adaptation and replay compete based on their respective reducible losses, determining what to replay and how much is needed.

%We further examine RoD’s candidate multiplier $m$, the primary selection hyperparameter in our implementation, which controls the number of candidates competing for each training batch. 
%Under our balanced candidate construction, $m<2$ implicitly imposes a minimum replay share as there are not enough adaptation candidates to fill the batch. At $m=1.5$, this constraint leads to consistently high replay, resulting in low forgetting but weaker adaptation (\cref{fig:ablation}b--c). Once $m\geq2$, adaptation candidates alone can fill the batch, and the replay allocation is determined entirely by joint selection. Increasing the candidate pool further provides additional selection freedom, but both the adaptation--forgetting trade-off and the resulting replay curriculum quickly stabilize, with similar behavior for $m=2$, $4$, and $8$. We therefore use $m=2$ throughout our main experiments, the smallest multiplier that leaves the replay allocation unconstrained.

We further examine RoD's candidate multiplier $m$, the primary selection hyperparameter in our implementation, which controls the number of candidates competing for each training batch. Under our balanced candidate construction, $m<2$ implicitly imposes a minimum replay share as there are not enough adaptation candidates to fill the batch. At $m=1.5$, this constraint leads to consistently high replay, resulting in low forgetting but weaker adaptation (\cref{fig:ablation}b--c). Once $m\geq2$, adaptation candidates alone can fill the batch, and the replay allocation is determined entirely by joint selection. Increasing the candidate pool further provides additional selection freedom, but both the adaptation--forgetting trade-off and the resulting replay curriculum quickly stabilize, with similar behavior for $m=2$, $4$, and $8$. We therefore use $m=2$ throughout our main experiments, the smallest multiplier that leaves the replay allocation unconstrained.

\begin{tcolorbox}[
  colback=white,
  colframe=takeawaysColor,
  boxrule=1.5pt,
  arc=3pt,
  left=4pt,right=4pt,top=4pt,bottom=4pt,
]
\textbf{\textcolor{takeawaysColor}{Takeaways.}}
%RoD benefits most when RHO-based selection can jointly determine which samples to replay and how much replay is needed, rather than operating within a fixed replay budget. Once the candidate pool allows this competition to operate without constraining the replay share, the resulting adaptation–forgetting trade-off remains stable across pool sizes.
Fixed-allocation baselines show that adaptive selection explains only part of RoD’s gains. Joint adaptation–replay competition yields the strongest trade-off by adapting both replay content and allocation to the model’s state. Once this replay allocation is unconstrained, the trade-off remains stable across candidate pool sizes.
\end{tcolorbox}

\subsection{RoD's model-dependent components transfer across model scales}
\label{sec:cross_scale}

% RoD relies on model-dependent signals for both its adaptation reference and online data selection. Applying these components at smaller scale could facilitate RoD for larger target models. Prior work provides evidence that data-selection signals can generalize across model scales~\citep{brandfonbrener2024color,khaddaj2025smalltolarge}. We test whether the same holds for RoD's adaptation reference and learned data curriculum. 
% In particular, transferring the curriculum would allow the model-dependent selection process to be performed at a smaller scale, while training the larger target model directly on the resulting curriculum.

RoD relies on model-dependent signals for both its adaptation reference and online data selection. Prior work provides evidence that data-selection signals can generalize across model scales~\citep{brandfonbrener2024color,khaddaj2025smalltolarge}. We therefore test whether RoD’s adaptation reference and learned data curriculum similarly transfer across scales, providing a route to perform its model-dependent computation on smaller models.

\paragraph{Can the adaptation reference be smaller?}
We run RoD on Qwen3.5-9B using the corresponding Qwen3.5-4B model as its adaptation reference. As shown in \cref{tab:cross_scale}, the smaller reference recovers the adaptation--forgetting trade-off of native RoD across adaptation loss, validation-loss forgetting, and task-based forgetting. Thus, the adaptation reference can operate below the target-model scale while recovering the resulting trade-off in this setting.

\paragraph{Can the curriculum be constructed at smaller scale?}

RoD’s online selection induces a data curriculum by determining which adaptation and replay samples enter each batch. We construct this curriculum with a smaller model, record the selected batches, and use the resulting sequence to train a larger model without RoD selection at scale (\cref{tab:cross_scale}).

% \begin{wraptable}{r}{0.50\textwidth}
% \vspace{-10pt}
% \centering
% \small
% \caption{\textbf{Cross-scale RoD.} Larger target models use native RoD or components constructed by a smaller model (\emph{ref.}: adaptation reference; \emph{curr.}: curriculum). Fixed replay provides a retention-matched reference.}
% \label{tab:cross_scale}
% \resizebox{\linewidth}{!}{
% \begin{tabular}{lccc}
% \toprule
% Method & Adapt. $\downarrow$ & Val. forget. $\downarrow$ & Task forget. $\downarrow$ \\
% \midrule
% \multicolumn{4}{l}{\textit{Qwen3.5-9B target (4B $\rightarrow$ 9B; \textbf{$2.25\times$})}} \\
% No-replay & 1.793 & 0.419 & 0.108 \\
% Fixed replay (32\%) & 1.807 & 0.019 & 0.039 \\
% \addlinespace[3pt]
% RoD (native) & 1.790 & 0.028 & 0.061 \\
% RoD (4B ref.) & 1.796 & 0.013 & 0.047 \\
% \addlinespace[3pt]
% RoD (4B curr.) & 1.795 & 0.019 & 0.052 \\
% \midrule
% \multicolumn{4}{l}{\textit{Nemotron-30B target (12B $\rightarrow$ 30B; \textbf{$2.5\times$})}} \\
% No-replay & 1.812 & 0.437 & 0.050 \\
% Fixed replay (32\%) & 1.818 & 0.130 & \placeholder{X} \\
% \addlinespace[3pt]
% RoD (12B curr.) & 1.805 & 0.076 & 0.027 \\
% \midrule
% \multicolumn{4}{l}{\textit{Qwen3.5-35B target (4B $\rightarrow$ 35B; \textbf{$8.75\times$})}} \\
% No-replay & 1.777 & 0.483 & 0.091 \\
% Fixed replay (32\%) & \placeholder{X} & \placeholder{X} & \placeholder{X} \\
% \addlinespace[3pt]
% RoD (4B curr.) & 1.751 & 0.024 & 0.050 \\
% \bottomrule
% \end{tabular}}
% \end{wraptable}

\begin{wraptable}{r}{0.55\textwidth}
\vspace{-10pt}
\centering
\small
\caption{\textbf{Cross-scale RoD.} Larger target models use native RoD or components constructed by a smaller model (\emph{ref.}: adaptation reference; \emph{curr.}: curriculum). No-replay and fixed replay provide reference points.}
\label{tab:cross_scale}
\resizebox{\linewidth}{!}{
\begin{tabular}{lccc}
\toprule
Method & Adapt. $\downarrow$ & Val. forget. $\downarrow$ & Task forget. $\downarrow$ \\
\midrule

\multicolumn{4}{l}{\textbf{\textit{Qwen3.5-9B target}} \textit{(4B $\rightarrow$ 9B; $2.25\times$)}} \\
No-replay & 1.793 & 0.419 & 0.108 \\
Fixed replay (32\%) & 1.807 & 0.019 & 0.039 \\
\cmidrule(lr){1-4}
RoD (native) & 1.790 & 0.028 & 0.061 \\
RoD (4B ref.) & 1.796 & 0.013 & 0.047 \\
RoD (4B curr.) & 1.795 & 0.019 & 0.052 \\

\midrule
\multicolumn{4}{l}{\textbf{\textit{Nemotron-30B target}} \textit{(12B $\rightarrow$ 30B; $2.5\times$)}} \\
No-replay & 1.812 & 0.437 & 0.050 \\
Fixed replay (32\%) & 1.818 & 0.130 & 0.023 \\
\cmidrule(lr){1-4}
RoD (12B curr.) & 1.805 & 0.076 & 0.027 \\

\midrule
\multicolumn{4}{l}{\textbf{\textit{Qwen3.5-35B target}} \textit{(4B $\rightarrow$ 35B; $8.75\times$)}} \\
No-replay & 1.777 & 0.483 & 0.091 \\
Fixed replay (32\%) & 1.753 & 0.027 & 0.044 \\
\cmidrule(lr){1-4}
RoD (4B curr.) & 1.751 & 0.024 & 0.050 \\

\bottomrule
\end{tabular}}
\vspace{-10pt}
\end{wraptable}

We first isolate this setting within Qwen3.5, where native 9B RoD provides a direct reference. Training Qwen3.5-9B on the curriculum constructed by Qwen3.5-4B recovers the adaptation--forgetting trade-off of native RoD despite a $2.25\times$ difference in model scale. The curriculum induced at 4B therefore captures learning and forgetting dynamics that remain informative for the 9B model.

We then increase both the target-model scale and the relative scale gap. We train Nemotron-30B on a curriculum constructed by Nemotron-12B ($2.5\times$) and Qwen3.5-35B on a curriculum constructed by Qwen3.5-4B ($8.75\times$). 
In both settings, the smaller-model curriculum retains the adaptation performance of no-replay CPT while substantially reducing forgetting. At the same time, it achieves an adaptation--retention trade-off competitive with the fixed 32\% replay baseline without specifying a replay ratio.
Overall, RoD's learned data curriculum generalizes across model scales, remaining effective even when the target model is $8.75\times$ larger than the model used to construct it.

\begin{tcolorbox}[
  colback=white,
  colframe=takeawaysColor,
  boxrule=1.5pt,
  arc=3pt,
  left=4pt,right=4pt,top=4pt,bottom=4pt,
]
\textbf{\textcolor{takeawaysColor}{Takeaways.}}
RoD’s model-dependent components transfer across model scales. A smaller adaptation reference recovers native RoD at 9B, while curricula constructed by smaller models retain competitive adaptation–forgetting trade-offs for target models up to 35B and scale gaps of up to $8.75\times$. This allows the model-dependent computation underlying RoD to be performed at substantially smaller scale than the target model.
\end{tcolorbox}

\section{Conclusion}

Continual pretraining must balance adapting to a new distribution with retaining capabilities acquired during pretraining. Replay can mitigate forgetting, but typically requires deciding in advance how much and what to replay, before the model's actual retention needs are known. We introduce \textbf{RoD}, which instead allows adaptation and replay examples to compete directly via complementary reducible-loss signals. Across domains, model families, and scales, RoD reaches or improves upon the adaptation--forgetting frontier of fixed replay without requiring a predefined replay ratio.

Our analyses show that RoD adapts replay to the model's evolving retention needs, targeting vulnerable sources and increasing replay as forgetting emerges. Ablations attribute these gains primarily to joint adaptation--replay competition rather than selection within either stream alone. Moreover, RoD's model-dependent computation can operate below the target-model scale: smaller adaptation references recover the trade-off of standard RoD, while transferred curricula remain effective across an 8.75$\times$ scale gap. 

Together, these results show that a strong stability–plasticity trade-off can emerge by replaying what is needed, when it is needed, rather than prescribing replay in advance. Our cross-scale results further provide practical paths for applying this principle to larger models.

\section*{Acknowledgements}

Lukas Thede thanks the International Max Planck Research School for Intelligent Systems (IMPRS-IS) for support. We are grateful for support by the Carl Zeiss Foundation, project "Certification and Foundations of Safe Machine Learning Systems in Healthcare". This work was partially funded by the ERC (853489 - DEXIM) and the Alfried Krupp von Bohlen und Halbach Foundation, which we thank for their generous support.

\section*{AI use statement}

We used generative AI tools to assist with code implementation, experimental verification, and the integration of author-developed mathematical material into the manuscript. Specifically, LLMs were used to implement individual functions following author-provided method specifications, improve code efficiency, check experimental configurations for consistency, sanity-check experiments and their outputs, and translate author-provided mathematical derivations and notes into manuscript-ready form. We did not use generative AI tools to formulate hypotheses, develop theoretical or conceptual frameworks, independently derive or prove mathematical claims, design the research methodology or experiments, or independently interpret results; the remaining required-disclosure tasks are not applicable to this work.

Additionally, we used generative AI tools to refine scientific figures based on author-specified changes and to improve the language, clarity, and consistency of author-written manuscript drafts. All AI-assisted code, experimental checks, mathematical content, figures, and text were reviewed and verified by the authors. We take responsibility for the final content of this work, including text, claims, code, and artifacts produced with the aid of generative AI.

\section*{Ethics statement}

This work studies methods for continued pretraining of language models and does not involve human subjects or the collection of personal or sensitive data. Our experiments use existing language models and datasets for research purposes. We do not identify ethical concerns specific to the proposed methodology beyond those generally associated with the development and adaptation of large language models. We have conducted this work in accordance with the ICLR Code of Ethics.

\section*{Reproducibility statement}

We facilitate reproducibility by building all experiments on publicly available datasets and open-weight language models. We describe RoD and its training procedure in \cref{sec:method}, with the complete algorithm and implementation details provided in \cref{app:method}. The experimental setup is summarized in \cref{sec:experiments} and documented in detail in \cref{app:experimental_setup}, including the models and datasets, preprocessing, optimization and training configuration, baselines, evaluation procedure, compute setup, and random seed. We additionally provide a code repository containing the implementation of RoD and the code required to reproduce the experiments and evaluations reported in this work: \url{https://github.com/bethgelab/replay-on-demand}.

\bibliography{iclr2027_conference}
\bibliographystyle{iclr2027_conference}

\newpage

\appendix

% ============================================================
\section{Additional Method Details}
\label{app:method}
% ============================================================

We provide additional motivation for RoD's source-specific reducible-loss scores and joint selection procedure, followed by the complete training algorithm and implementation details.

% ------------------------------------------------------------
\subsection{Deriving Source-Specific Reducible Loss from RHO}
\label{app:rho_derivation}
% ------------------------------------------------------------

RoD uses different reference models for adaptation and replay: an adaptation specialist $\theta_{\mathrm{ref}}^A$ estimates what remains to be learned on the adaptation distribution, while the pretrained model $\theta_0$ measures degradation on replay data. We motivate this construction from the original RHO objective.

Let
\begin{equation}
L(x\mid\mathcal{S})
=
-\frac{1}{n}\log p(x\mid\mathcal{S})
\end{equation}
denote the per-token loss under the posterior predictive of a Bayesian learner that has observed data $\mathcal{S}$, where $n=4096$ is the number of loss-bearing tokens per candidate. During CPT, the learner has observed
\begin{equation}
    \mathcal{S}_t = \mathcal{D}_0 \cup B_{1:t},
\end{equation}
consisting of the pretraining corpus $\mathcal{D}_0$ underlying $\theta_0$ and the CPT batches $B_{1:t}$ encountered so far.

Following the derivation of RHO~\citep{mindermann2022prioritized}, consider a holdout
$\mathcal{H}=\mathcal{H}_A\cup\mathcal{H}_R$
containing data from both the adaptation and replay distributions. The utility of a candidate $x$ can be expressed through the improvement it induces in the likelihood of this holdout. By Bayes' rule,
\begin{equation}
\log p(\mathcal{H}\mid\mathcal{S}_t,x)
-
\log p(\mathcal{H}\mid\mathcal{S}_t)
=
n\big[
L(x\mid\mathcal{S}_t)
-
L(x\mid\mathcal{S}_t,\mathcal{H})
\big].
\label{eq:rho_bayes}
\end{equation}
Thus, candidate utility is determined by the difference between its loss under the current learner and the loss attainable after conditioning on the holdout.

To obtain the RoD scores used in practice, we make three approximations. First, as in RHO, we approximate the current posterior predictive with the current model,
\begin{equation}
    L(x\mid\mathcal{S}_t)
    \approx
    \ell_{\theta_t}(x).
\end{equation}
Second, we approximate the posterior conditioned on the holdout by dropping the particular CPT trajectory while retaining the original pretraining data,
\begin{equation}
    L(x\mid\mathcal{S}_t,\mathcal{H})
    \approx
    L(x\mid\mathcal{D}_0,\mathcal{H}).
\end{equation}
Finally, we assume that the relevant component of the holdout depends on the source of the candidate:
\begin{equation}
L(x\mid\mathcal{D}_0,\mathcal{H}_A,\mathcal{H}_R)
\approx
\begin{cases}
L(x\mid\mathcal{D}_0,\mathcal{H}_A)
\approx
\ell_{\theta_{\mathrm{ref}}^A}(x),
& x\in C_A,\\[2pt]
L(x\mid\mathcal{D}_0)
\approx
\ell_{\theta_0}(x),
& x\in C_R.
\end{cases}
\label{eq:il_factorization}
\end{equation}
For replay examples, $\mathcal{D}_0$ already contains substantially more data from the replay distribution than the additional holdout $\mathcal{H}_R$, while adaptation data provide comparatively little information about $\mathcal{D}_R$. The pretrained model therefore provides the natural irreducible-loss reference for replay. For adaptation examples, $\theta_{\mathrm{ref}}^A$ instead approximates the model obtained after learning the adaptation distribution. Importantly, its forgetting on the pretraining distribution does not enter the score because the adaptation reference is evaluated only on $\mathcal{D}_A$.

Together, these approximations recover the source-specific scores used by RoD,
\begin{equation}
\rho(x;\theta_t)
=
\begin{cases}
    \rho_A(x;\theta_t)
    = \ell_{\theta_t}(x) - \ell_{\theta_{\mathrm{ref}}^A}(x),
    & x \in C_A, \\
    \rho_R(x;\theta_t)
    = \ell_{\theta_t}(x) - \ell_{\theta_0}(x),
    & x \in C_R.
\end{cases}
\label{eq:rod_derivation}
\end{equation}
This derivation provides a common interpretation of both scores as excess loss relative to a source-appropriate reference.

\paragraph{Interpretation under exact Bayesian updating.}
Under exact Bayesian inference, sequential updating recovers the joint posterior~\citep{farquhar2019unifying}. For replay examples, the dominance of the original pretraining data,
$|\mathcal{D}_0|\gg|B_{1:t}|$,
then implies
\begin{equation}
L(x\mid\mathcal{S}_t)
\approx
L(x\mid\mathcal{D}_0)
\approx
L(x\mid\mathcal{S}_t,\mathcal{H}),
\qquad
x\sim\mathcal{D}_R.
\end{equation}
The corresponding ideal replay score is therefore approximately zero. Under the approximation
$L(x\mid\mathcal{S}_t)\approx\ell_{\theta_t}(x)$,
the practical replay score $\rho_R$ can consequently be interpreted as measuring the per-example deviation of the continually trained model from this idealized sequential learner.

\paragraph{Why use source-specific references?}
Applying standard RHO directly would require a single reference trained on
$\mathcal{H}_A\cup\mathcal{H}_R$.
In CPT, such a model would itself correspond to one particular adaptation--retention trade-off, determined by the mixture of adaptation and replay data used to construct it. The desired posterior conditioned on both $\mathcal{D}_0$ and adaptation data therefore cannot generally be represented by a single SGD-trained reference without first choosing how to balance these objectives.

RoD instead approximates this reference separately for the two candidate sources. When the source-matched reference is the better of the two references for its corresponding examples, the resulting score coincides, up to at most
$(\log 2)/n\approx1.7\times10^{-4}$
nats per token, with RHO using an equal-weight Bayesian model average of the two references:
\begin{equation}
\rho(x)
\approx
\ell_{\theta_t}(x)
-
\min\{
    \ell_{\theta_{\mathrm{ref}}^A}(x),
    \ell_{\theta_0}(x)
\}.
\end{equation}
This provides an alternative interpretation in which the two source-specific references approximate a common reference model while avoiding the need to solve the adaptation--retention trade-off during reference construction.

\paragraph{Why replay requires online scoring.}
The distinction between adaptation and replay also explains why the replay signal must depend on the current model. Replacing $\theta_t$ with $\theta_0$ in the adaptation score gives
\begin{equation}
    \ell_{\theta_0}(x)
    -
    \ell_{\theta_{\mathrm{ref}}^A}(x),
\end{equation}
which recovers the offline conditional loss-reduction signal used by CoLoR-Filter~\citep{brandfonbrener2024color}. Applying the same substitution to replay instead gives
\begin{equation}
    \ell_{\theta_0}(x)-\ell_{\theta_0}(x)=0.
\end{equation}
Unlike adaptation potential, forgetting is therefore not a static property of an example. It arises along the model's training trajectory and must be measured online.

% ------------------------------------------------------------
\subsection{Joint Selection as Resource Allocation}
\label{app:joint_selection_theory}
% ------------------------------------------------------------

The previous derivation motivates the source-specific scores. We next provide an interpretation of why comparing these scores through joint top-$k$ selection yields an adaptive allocation between adaptation and replay.

Both $\rho_A$ and $\rho_R$ measure excess code length, in nats per token, relative to the corresponding source-specific reference. For a relative retention weight $\kappa\geq0$, consider
\begin{align}
J_\kappa(\theta)
&=
\E_{x\sim\mathcal{D}_A}
\big[\rho_A(x;\theta)\big]
+
\kappa\,
\E_{x\sim\mathcal{D}_R}
\big[\rho_R(x;\theta)\big]
\nonumber\\
&=
\tfrac{1}{n}
\big[
\KL(\mathcal{D}_A\|p_\theta)
-
\KL(\mathcal{D}_A\|p_{\theta_{\mathrm{ref}}^A})
\big]
+
\tfrac{\kappa}{n}
\big[
\KL(\mathcal{D}_R\|p_\theta)
-
\KL(\mathcal{D}_R\|p_{\theta_0})
\big].
\label{eq:rod_objective}
\end{align}
This objective makes explicit the exchange rate between reducing adaptation loss and preserving the replay distribution.

For native replay, where $\mathcal{D}_0\sim\mathcal{D}_R$, and under a flat prior, the negative log-posterior of Bayesian CPT on $\mathcal{D}_0$ and the adaptation corpus equals $N_A$ times the empirical $J_\kappa$ up to a constant, with
\begin{equation}
    \kappa=\frac{N_0}{N_A},
\end{equation}
where $N_0$ and $N_A$ denote the respective token counts. For Nemotron,
$N_0\approx2\times10^{13}$~\citep{nvidia2025nvidianemotronnano2}
and
$N_A\approx5\times10^{9}$
(\cref{app:data}), giving
$\kappa\approx4\times10^3$.
A fixed replay mixture with replay share $r$ instead corresponds at convergence to
\begin{equation}
    \kappa=\frac{r}{1-r}.
\end{equation}
Practical CPT therefore deliberately chooses an operating point that trades some retention of the full pretraining posterior for stronger adaptation.

The replay component also connects to parameter-space regularization. For native replay,
\begin{equation}
\nabla_\theta
\E_{\mathcal{D}_R}[\ell_\theta]
\big|_{\theta_0}
\approx 0.
\end{equation}
Letting $\Delta=\theta-\theta_0$ and using the Fisher approximation $F_R$ to the Hessian, valid when $p_{\theta_0}\approx\mathcal{D}_R$, gives
\begin{equation}
\E_{\mathcal{D}_R}[\rho_R]
\approx
\frac{1}{2}\Delta^\top F_R\Delta.
\end{equation}
This corresponds to the full-Fisher form of the EWC penalty~\citep{Kirkpatrick_2017,schwarz2018progresscompressscalable}. RoD targets the same local notion of deviation from the pretrained model through data allocation rather than by adding an explicit parameter-space regularizer.

\paragraph{Marginal utility of a training example.}
We can further relate the reducible-loss score to the expected utility of assigning a training slot to an example. Suppose that, for examples with $\rho(x)>0$, the loss satisfies a per-example Polyak--{\L}ojasiewicz-type condition relative to the reference loss,
\begin{equation}
    \|\nabla\ell_\theta(x)\|^2
    \ge
    2\mu_s\rho(x),
    \qquad
    s=s(x)\in\{A,R\},
\end{equation}
and that $\ell$ is $\beta$-smooth. A gradient step on $x$ with $\eta\le1/\beta$ then decreases its loss by at least
\begin{equation}
    \eta\mu_s\rho(x).
\end{equation}
Treating this bound as tight, neglecting interactions between examples as in RHO batch selection, and assuming that the reduction transfers to the corresponding source distribution, the gain from assigning one batch slot to $x$ can be approximated as
\begin{equation}
    g(x)
    \approx
    \eta\,w_s\mu_s\,\rho(x),
\end{equation}
where $w_A=1$ and $w_R=\kappa$.

Under these assumptions, maximizing
\begin{equation}
    \sum_{x\in B_t}g(x)
    \qquad
    \text{subject to}
    \qquad
    |B_t|=k
\end{equation}
is separable across candidates. Selecting the top-$k$ examples therefore maximizes this local surrogate exactly. For
\begin{equation}
    \kappa=\frac{\mu_A}{\mu_R},
\end{equation}
the source-dependent factors cancel and the selection reduces to the unweighted RoD rule,
\begin{equation}
    B_t
    =
    \operatorname{TopK}_{x\in C_A\cup C_R}
    \rho(x;\theta_t).
\end{equation}

This view provides a resource-allocation interpretation of RoD. The $k$-th largest score $\tau_t$ acts as the current shadow price of a training slot: examples from each source are selected until their marginal reducible loss falls below this threshold. The replay share is therefore not prescribed independently, but emerges from the relative distributions of adaptation and replay scores at each training step. RoD uses the unweighted comparison as a parameter-free default, placing one nat of adaptation reducible loss and one nat of replay reducible loss on equal footing.

% ------------------------------------------------------------
\subsection{Interpretation under Proxy Replay}
\label{app:proxy_replay_theory}
% ------------------------------------------------------------

The preceding interpretation is cleanest when replay examples originate from the model's own pretraining distribution. When only proxy replay data are available, as in our Qwen3.5 experiments, the pretrained model need not be converged on the replay distribution. This changes the interpretation of the replay score.

Let $q$ denote the proxy replay distribution. If
\begin{equation}
    \bar g_q
    =
    \nabla_\theta
    \E_q[\ell_\theta]
    \big|_{\theta_0}
    \neq 0,
\end{equation}
and $H_q$ denotes the corresponding Hessian, then
\begin{equation}
\E_{x\sim q}[\rho_R(x;\theta)]
=
\tfrac{1}{n}
\big[
\KL(q\|p_\theta)
-
\KL(q\|p_{\theta_0})
\big]
\approx
\bar g_q^\top\Delta
+
\tfrac{1}{2}\Delta^\top H_q\Delta.
\label{eq:proxy_replay_expansion}
\end{equation}
Unlike the native-replay case, the linear term does not vanish. The model can therefore reduce its replay score partly by fitting the proxy distribution $q$, rather than exclusively by returning toward the pretrained model $\theta_0$. This provides one explanation for why validation-loss retention on proxy replay can be less tightly coupled to task-based retention, as observed for Qwen3.5 in \cref{app:qwen_proxy_replay}.

One alternative would be to measure functional deviation from the pretrained model directly through self-distillation,
\begin{equation}
\rho_R^{\mathrm{KD}}(x)
=
\frac{1}{n}
\sum_j
\KL\big(
p_{\theta_0}(\cdot\mid x_{<j})
\,\|\,
p_{\theta_t}(\cdot\mid x_{<j})
\big)
\ge 0.
\end{equation}
This score is minimized at $\theta_0$ for any replay distribution $q$ and would therefore remove the linear proxy-fitting term, in the spirit of distillation-based replay and functional regularization~\citep{rolnick2019experience,buzzega2020dark,titsias2019functional}. We do not use this variant in our experiments because it requires storing or recomputing the pretrained model's token-level predictive distributions on replay data, whereas our loss-based reference scores can be precomputed as a single scalar per sequence.

% ------------------------------------------------------------
\subsection{RoD Algorithm and Implementation}
\label{app:algorithm}
% ------------------------------------------------------------

We provide the complete RoD training procedure in \cref{alg:rod} and detail the candidate construction, loss computation, and reference-loss caching used in our implementation.

\paragraph{Candidate construction.}
Before training, both the adaptation corpus $\mathcal{D}_A$ and replay corpus $\mathcal{D}_R$ are tokenized and packed into fixed-length blocks of 4096 tokens. Documents are packed only within the same data source, with an end-of-document token inserted between concatenated documents. Each resulting block therefore belongs unambiguously to either the adaptation or replay distribution.

At each training step, RoD constructs a candidate pool of $mk$ sequences, where $k$ is the trained batch size and $m$ the candidate multiplier. We balance this pool between adaptation and replay by subsampling the replay corpus once before training such that both sources occur with equal probability in the shuffled candidate stream. Consequently, $|C_A|=|C_R|=mk/2$ in expectation, independent of the relative sizes of the original corpora. Under this construction, $m=2$ is the smallest multiplier for which adaptation candidates alone can fill the complete training batch; for $m<2$, a minimum replay share is imposed by construction. We therefore use $m=2$ by default and study this choice in \cref{sec:ablations}.

\paragraph{Sequence-level loss and reference caching.}
We score each candidate using its token-normalized, full-sequence language-modeling loss
\begin{equation}
    \ell_\theta(x)
    =
    \frac{1}{|T(x)|}
    \sum_{j\in T(x)}
    -\log p_\theta(x_j\mid x_{<j}),
    \label{eq:sequence_loss}
\end{equation}
where $T(x)$ denotes the loss-bearing token positions in sequence $x$. All tokens in our packed pretraining sequences are loss-bearing. The same aggregation is used for the target and reference models, ensuring that their loss differences are directly comparable.

Because both reference models remain fixed throughout training, their losses are precomputed once. The adaptation specialist $\theta_{\mathrm{ref}}^A$ scores every adaptation block, while the pretrained model $\theta_0$ scores every replay block. Each loss is stored together with a deterministic hash of the block's source and token IDs. During training, candidate sequences are matched to their cached reference losses through this hash; neither reference model therefore requires an online forward pass.

\paragraph{Online scoring and joint selection.}
At each step $t$, the current model $\theta_t$ first performs a forward-only pass over the complete candidate pool $C_A\cup C_R$. We combine the resulting current-model losses with the cached reference losses to compute
\begin{equation}
\rho(x;\theta_t)
=
\begin{cases}
    \ell_{\theta_t}(x)-\ell_{\theta_{\mathrm{ref}}^A}(x),
        & x\in C_A, \\[2pt]
    \ell_{\theta_t}(x)-\ell_{\theta_0}(x),
        & x\in C_R.
\end{cases}
\end{equation}
RoD then jointly ranks all $mk$ candidates and selects the $k$ highest-scoring sequences,
\begin{equation}
    B_t =
    \operatorname{TopK}_{x\in C_A\cup C_R}
    \rho(x;\theta_t).
\end{equation}
No source-specific quota is applied: adaptation and replay candidates compete directly for the same $k$ training slots. Padding introduced for batch-size divisibility is assigned a score of $-\infty$ and cannot enter the selected batch.

Finally, we perform a standard forward--backward pass on $B_t$ and update the model using the unweighted autoregressive language-modeling objective. RHO therefore affects training only through data selection; selected examples are not reweighted according to their scores. Each RoD step consequently consists of one forward-only scoring pass over $mk$ candidates, a cached reference-loss lookup and joint top-$k$ selection, followed by one forward--backward pass over the selected $k$ examples. All reference-model computation is performed offline.

\begin{algorithm}[t]
\caption{RoD: joint reducible-loss selection for continued pretraining.}
\label{alg:rod}
\begin{algorithmic}[1]
\Require adaptation data $\mathcal{D}_A$, replay data $\mathcal{D}_R$,
adaptation reference $\theta_{\mathrm{ref}}^A$, pretrained model $\theta_0$,
batch size $k$, candidate multiplier $m$
\State Initialize target model $\theta \leftarrow \theta_0$
\State Precompute $\ell_{\theta_{\mathrm{ref}}^A}(x)$ for all $x\in\mathcal{D}_A$
\State Precompute $\ell_{\theta_0}(x)$ for all $x\in\mathcal{D}_R$
\For{each training step}
    \State Draw $mk$ candidates from the balanced mixture of
           $\mathcal{D}_A$ and $\mathcal{D}_R$, yielding $C_A$ and $C_R$
    \State Compute $\ell_\theta(x)$ for all $x\in C_A\cup C_R$
           \Comment{forward only}
    \For{$x\in C_A\cup C_R$}
        \If{$x\in C_A$}
            \State $\rho(x)\leftarrow
            \ell_\theta(x)-\ell_{\theta_{\mathrm{ref}}^A}(x)$
        \Else
            \State $\rho(x)\leftarrow
            \ell_\theta(x)-\ell_{\theta_0}(x)$
        \EndIf
    \EndFor
    \State $B\leftarrow
    \operatorname{TopK}_{x\in C_A\cup C_R}\rho(x)$
           \Comment{$|B|=k$}
    \State Update $\theta$ on $B$ using the standard language-modeling loss
\EndFor
\end{algorithmic}
\end{algorithm}

% ------------------------------------------------------------
\subsection{Adaptation Reference Construction}
\label{app:adaptation_reference}
% ------------------------------------------------------------

For each model--domain setting, we obtain the adaptation reference $\theta_{\mathrm{ref}}^A$ by continuing pretraining of the corresponding base model exclusively on the adaptation corpus, without replay or other forgetting mitigation. We first train with a warmup followed by a constant learning rate until the adaptation validation loss converges, which occurs after approximately two epochs across our settings. Once convergence is reached, we append an additional $0.5$ epochs of cosine learning-rate decay and use the resulting checkpoint as $\theta_{\mathrm{ref}}^A$. The same model constitutes the no-replay CPT baseline in our experiments.

These adaptation specialists achieve the strongest, or comparable-to-strongest, adaptation performance among our training runs, but exhibit substantial forgetting of the pretrained distribution. This forgetting is inconsequential for their role in RoD: the adaptation reference is not intended to provide a desirable adaptation--retention trade-off, but to estimate the loss attainable after learning the adaptation distribution. Its loss therefore provides a target against which RoD measures the remaining learning potential of each adaptation example. Moreover, this estimate need not be produced at the target-model scale: in \cref{sec:cross_scale}, we show that a 4B adaptation reference recovers the adaptation--forgetting trade-off of the standard 9B reference within the Qwen3.5 family.

% ============================================================
\section{Extended Experimental Setup}
\label{app:experimental_setup}

We provide additional details on the models, data, training setup, baselines, and evaluation used in our experiments.
Our experimental settings are designed to progressively test RoD beyond its primary Nemotron-12B setting.
We first evaluate Nemotron-12B on both Legal and German adaptation, where the original pretraining distribution is available for replay.
We then evaluate Qwen3.5-9B on the same two domains to test whether RoD generalizes across model families and remains effective when the original pretraining data are unavailable and replay must instead rely on a proxy distribution.
Qwen3.5-4B provides an additional smaller-scale setting within the Qwen family and, more importantly, serves as the source model for our cross-scale curriculum-transfer experiments; we therefore evaluate it only on German, the domain used for these scaling experiments.
Finally, we test scalability to substantially larger models by transferring RoD curricula constructed at smaller scale to Nemotron-30B-A3B and Qwen3.5-35B-A3B, avoiding online curriculum construction at the target-model scale.
Together, these settings separate generalization across domains and model families from scalability through curriculum transfer.

Unless stated otherwise, all methods within a model--domain setting use the same preprocessing, optimization setup, and trained-token budget.
All experiments use a sequence length of 4096 tokens and a trained batch of 1024 sequences, corresponding to approximately 4.19M trained tokens per update.

\subsection{Models}
\label{app:models}

We use Nemotron-Nano-12B-v2 and Qwen3.5-9B for the primary Legal and German experiments, with Qwen3.5-4B additionally evaluated on German.
For the cross-scale curriculum-transfer experiments in \cref{sec:cross_scale}, we use Nemotron-3-Nano-30B-A3B and Qwen3.5-35B-A3B as larger target models.

Nemotron-Nano-12B-v2 \citep{nvidia2025nvidianemotronnano2} is a 12B-parameter hybrid architecture combining Mamba-style state-space layers with periodic attention.
For the large-scale transfer experiment, we additionally use Nemotron-3-Nano-30B-A3B \citep{nvidia2026nemotron3ultraopen}, a 30B-parameter mixture-of-experts model with approximately 3B active parameters per token.
Qwen3.5-9B and Qwen3.5-4B \citep{qwen3.5} are dense hybrid models combining Gated DeltaNet layers with periodic full attention.
We additionally use Qwen3.5-35B-A3B \citep{qwen3.5} as the target of our largest curriculum-transfer experiment.
Although the Qwen3.5 checkpoints use a multimodal model class, all continued-pretraining experiments are text-only.
We use each model's corresponding tokenizer, with the 4B and 9B Qwen3.5 models sharing the same tokenizer.

\subsection{Adaptation and Replay Data}
\label{app:data}

\paragraph{Adaptation data.}
For Legal adaptation, we use the Nemotron-Pretraining-Legal-v1 corpus \citep{nvidia2026nemotron3ultraopen}. The corpus contains legal text from multiple sources, including case law, regulatory text, and legal question-answering data, and is dominated by CaseHold, which accounts for approximately 82\% of the adaptation corpus. For German adaptation, we construct a German-language corpus following the data collection of \citet{soofiteam2026sovereignopensourcefoundationmodel}. The corpus comprises nine sources spanning German PDF and Wikipedia data as well as cultural, economic, legal, news, political, scientific, and web data.

\paragraph{Replay data.}
For replay, we use the general Nemotron pretraining corpus \citep{nvidia2025nvidianemotronnano2}. It contains data from 14 fine-grained sources spanning world knowledge, reasoning and mathematics, multilingual data, and code. For Nemotron, these data originate from its own pretraining distribution and therefore provide a direct approximation to the knowledge acquired during pretraining. The original Qwen3.5 pretraining data are not publicly available. We therefore use the same Nemotron general corpus as a proxy for replay in Qwen3.5. As discussed in \cref{app:replay_distribution,app:qwen_proxy_replay}, this distinction is important because loss degradation on proxy replay need not be as closely aligned with capability forgetting as degradation on the model's own pretraining distribution.

\paragraph{Preprocessing and splits.}
We tokenize all corpora using the respective target model's tokenizer and pack the documents into sequences of 4096 tokens. Packing is performed within each source so that every sequence retains a unique source label for our source-level analyses. Partial final sequences are discarded. We associate every packed sequence with a content hash, which allows us to consistently match examples between reference-loss precomputation, training, and evaluation.
We construct a deterministic 1\% validation holdout set independently for each source, using the content hash and seed 42. These examples are excluded from training and form the source-specific validation sets used to measure adaptation and forgetting. For German, we subsample the adaptation corpus to 25\% of its documents and cap individual sources to balance coverage across the nine constituent sources. Replay data are subsampled to 20\% of the available documents for both domains.
After preprocessing and subsampling, the resulting Legal and German corpora contain approximately 9.6B and 14.2B tokens, respectively.
For Legal, adaptation and replay data form an approximately balanced 50:50 mixture, while the German corpus contains approximately 35\% adaptation and 65\% replay data.
For RoD, we balance the candidate stream between adaptation and replay before joint selection. Consequently, the available corpus proportions do not impose a minimum replay share on RoD; the selected replay fraction is determined by the joint RHO ranking.

\subsection{Training Setup}
\label{app:training_setup}

\paragraph{Optimization.}
We train all continued-pretraining models using Adam with $\beta_1=0.9$, $\beta_2=0.95$, $\epsilon=10^{-8}$, weight decay $0.1$, and gradient clipping at $1.0$.
Training uses bfloat16 precision and a distributed optimizer.
We select the learning rate through preliminary adaptation experiments rather than tuning it separately for RoD.
An initial learning-rate pilot at smaller batch scale identified $1\times10^{-5}$ as the strongest setting among the evaluated values.
We subsequently scale and tune the learning rate for the full training configuration, resulting in a peak learning rate of $1.2\times10^{-4}$, which we keep fixed across all primary model--domain settings.
The minimum learning rate is $1.2\times10^{-5}$.

We use a warmup--stable--decay schedule.
After a short warm-up, training proceeds at a constant learning rate until the adaptation validation loss of the corresponding no-replay baseline converges, followed by a cosine decay to the minimum learning rate.
The resulting trained-token-matched budgets range from 2,824 to 2,980 updates across the primary model--domain settings, corresponding to approximately 11.8--12.5B trained tokens.
Within each setting, RoD and all trained-token-matched fixed-replay baselines use the same optimization schedule and trained-token budget.

\paragraph{Batch construction.}
Each training update contains 1024 sequences of length 4096, corresponding to 4,194,304 trained tokens.
For RoD, the candidate multiplier $m$ controls the number of sequences considered for selection while the trained batch remains fixed.
We use $m=2$ throughout the main experiments and vary only $m$ in the candidate-multiplier ablation (\Cref{sec:ablations}).
The construction and selection of RoD candidates are described in \Cref{app:method}.

\paragraph{Cross-scale curriculum transfer.}
For the cross-scale experiments in \cref{sec:cross_scale}, we construct the RoD curriculum using a smaller source model and reuse it to train a larger target model without performing online selection at the target scale.
Specifically, we record the training batches selected by RoD throughout the source-model run, preserving both their adaptation--replay composition and ordering, and train the target model on this fixed sequence of examples.
The resulting curriculum therefore transfers RoD's learned allocation of what to train on and when, while avoiding the model-dependent reference-loss computation and candidate selection at the larger target scale.

\paragraph{Runs and compute.}
We use seed 42 throughout training and run each experimental condition once.
Accordingly, the reported frontiers are single-seed estimates.
We assess the consistency of our findings by replicating across model families, model scales, and adaptation domains, rather than by repeated runs of individual settings.
Small differences between nearby operating points should therefore not be interpreted as statistically significant.
Our conclusions instead rely on qualitative patterns that recur across model families, scales, adaptation domains, and replay settings.
Training is distributed across NVIDIA B200 GPUs, with the number of GPUs adjusted to the model and experimental setting.
Reference losses are computed separately before RoD training using forward-only evaluation and cached by content hash.
Thus, the reference models introduce an offline computation cost but do not need to be evaluated during RoD training.

\subsection{Baselines}
\label{app:baselines}

We compare RoD against three strategies for navigating the adaptation--retention trade-off: continued pretraining without replay, continued pretraining with fixed replay allocations, and post-hoc model merging.

\paragraph{No-replay CPT.}
The no-replay baseline continues pretraining exclusively on the adaptation corpus without replay.
It therefore represents the adaptation-focused endpoint of the replay frontier and exposes the forgetting that arises in the absence of explicit retention measures.
The same training procedure is used to construct the domain specialist that serves as RoD's adaptation reference.

\paragraph{Fixed-replay CPT.}
Fixed-replay baselines mix adaptation and replay examples at a constant ratio throughout training without data selection.
We construct three replay settings by retaining 0.11, 0.25, and 1.0 of the available replay corpus while keeping the adaptation corpus fixed.
Because the available adaptation--replay corpus ratios differ between domains, these same subsampling configurations result in different realized training mixtures.
For Legal, they correspond to approximately $10\%$, $20\%$, and $50\%$ replay; for German, they correspond to approximately $17\%$, $32\%$, and $65\%$ replay.

For our primary comparison, we train each fixed-replay baseline with the same trained-token budget as the no-replay CPT baseline.
Sweeping the replay allocation provides an oracle-like comparison that approximates the operating points available when the replay ratio can be selected in hindsight, whereas RoD determines its replay allocation within a single run.
We additionally report a \emph{max-replay} reference trained on all available adaptation and replay data.
These runs use approximately twice the trained-token budget and are therefore treated as higher-budget references rather than trained-token-matched baselines.
\Cref{app:extended_frontier} further examines this distinction by continuing the German fixed-replay baselines beyond the matched trained-token budget until convergence.

\paragraph{Model merging.}
As a post-hoc alternative to replay, we linearly interpolate the parameters of the pretrained base model $\theta_0$ and the no-replay model $\theta_A$, following the weight-space interpolation used in model soups~\citep{wortsman2022modelsoups},
\begin{equation}
    \theta_{\mathrm{merge}}
    =
    (1-\lambda)\theta_0 + \lambda\theta_A,
\end{equation}
with $\lambda\in\{0.2,0.4,0.6,0.8\}$.
Merging is performed independently for every floating-point parameter and requires no additional training.
Varying $\lambda$ yields a post-hoc adaptation--retention trade-off between the pretrained and adapted models.

Across training-based methods, we keep the optimizer, learning-rate schedule, sequence length, trained batch size, seed, and matched trained-token budget fixed within each model--domain setting.
The primary difference, therefore, lies in how the training budget is allocated between adaptation and replay.

\subsection{Evaluation}
\label{app:evaluation}

We evaluate continual pretraining along two complementary dimensions: adaptation to the target distribution and retention of capabilities acquired during pretraining.

\paragraph{Validation-loss evaluation.}
We measure adaptation using language-modeling loss on held-out data from the adaptation corpus.
Our objective is specifically to model the target distribution used for continued pretraining, rather than to optimize a predefined set of downstream Legal or German capabilities.
Held-out language-modeling loss therefore directly measures how well the model has learned this distribution without introducing assumptions about which downstream tasks should benefit from adaptation.
We compute loss separately for every adaptation source and aggregate sources according to their corpus prevalence, such that lower adaptation loss indicates stronger modeling of the target distribution.

We analogously evaluate language-modeling loss on held-out general data to measure forgetting at the distribution level.
For each source $o$, we compare the continually pretrained model's loss $\ell_{\theta}(o)$ with that of the pretrained base model $\ell_{\theta_0}(o)$.
We define validation-loss forgetting as
\begin{equation}
    F_{\mathrm{val}}(\theta)
    =
    \sum_{o\in\mathcal{D}_R}
    w_o
    \max\!\left(0,\,
        \ell_{\theta}(o)-\ell_{\theta_0}(o)
    \right),
\end{equation}
where $w_o$ denotes the source's corpus weight.
Capping source-level differences at zero prevents improvements on one source from compensating for degradation on another.
Both the base model and all continually pretrained checkpoints are evaluated on the same source-specific holdouts.
We use the corpus-weighted metric throughout the main paper and additionally retain unweighted source averages as a breadth-oriented diagnostic.

This distribution-level metric is conceptually aligned with RoD’s replay signal, as both quantify degradation in language-modeling loss relative to the pretrained model. Importantly, forgetting is evaluated on held-out data excluded from training, testing whether RoD’s example-level replay selection translates into retention beyond the selected training examples.

\paragraph{Task-based capability evaluation.}
Validation loss measures retention directly on the replay distribution, but changes in language-modeling loss do not necessarily translate into changes in downstream capabilities.
We therefore complement it with a task-based evaluation organized according to the capability taxonomy introduced by CapTrack~\citep{thede2026captrackmultifacetedevaluationforgetting}.
Because we evaluate base models without post-training, we focus on latent capabilities that can be meaningfully assessed in pretrained models.
Specifically, we evaluate four capability groups: \emph{parametric knowledge}, \emph{reasoning and problem solving}, \emph{commonsense and robustness}, and \emph{multilingual capabilities}.
\Cref{tab:capability_evals} lists the benchmarks used to instantiate each capability.

\begin{table}[t]
    \centering
    \small
    \caption{\textbf{Task-based capability evaluation.}
    Benchmarks used to evaluate the four pretraining-time capability groups considered in our experiments.}
    \label{tab:capability_evals}
    \begin{tabular}{p{0.23\linewidth}p{0.71\linewidth}}
        \toprule
        Capability & Evaluation tasks \\
        \midrule
        Parametric knowledge &
        MMLU World Religions, Prehistory, High School Geography,
        High School Government and Politics, US Foreign Policy,
        Security Studies, Global Facts, Miscellaneous,
        Clinical Knowledge, Medical Genetics, Professional Medicine,
        Anatomy, Management, Marketing, Business Ethics, and
        Professional Accounting; SciQ; OpenBookQA; ARC-Easy;
        MedQA; MMLU-Pro. \\
        \addlinespace
        Reasoning and problem solving &
        BBH; MuSR; MMLU Formal Logic, Logical Fallacies, Econometrics,
        High School Microeconomics, High School Macroeconomics,
        High School Mathematics, College Mathematics,
        Elementary Mathematics, Abstract Algebra, and
        High School Statistics; ARC-Challenge; AQuA-RAT; SAT Math. \\
        \addlinespace
        Commonsense and robustness &
        HellaSwag; PIQA; WinoGrande; CommonsenseQA. \\
        \addlinespace
        Multilingual capabilities &
        multilingual MMLU in German, French, Spanish, Russian,
        Chinese, and Arabic; BeleBele in German, French, Spanish,
        Russian, Chinese, Arabic, and English. \\
        \bottomrule
    \end{tabular}
\end{table}

All tasks are evaluated as multiple-choice problems using our evaluation suite with a vLLM backend.
Following the underlying benchmark configurations, we report length-normalized accuracy for tasks that define it and raw accuracy otherwise.
Base and continually pretrained models are evaluated using identical task configurations and prompting setups.

For each capability, we aggregate performance across its constituent tasks and compute the decrease relative to the corresponding base model, capped at zero.
Task-based forgetting is then defined as the mean forgetting across the four capability groups.
As for validation-loss forgetting, this construction prevents improvements in one capability from compensating for degradation in another.

% ============================================================
\section{Extended Experiments and Analyses}
\label{app:extended_experiments}
% ============================================================

\subsection{Additional Adaptation--Forgetting Results}
\label{app:extended_frontier}

\begin{table}[t]
\centering
\small
\caption{\textbf{Full adaptation--forgetting frontier, all settings.} Adaptation loss, general validation-loss forgetting, and task-based forgetting (capped mean accuracy drop over four capabilities) for every arm plotted in \cref{fig:frontier}. All arms within a setting are compute-matched (same optimizer steps from base) except \emph{max-replay}, trained to its own epoch target at $\approx 2\times$ the tokens. Replay share is the realized fraction of the trained batch for fixed-replay arms and the emergent fraction for RoD; model merging has no replay share (weight-space interpolation). Val.\ forget.\ and task forget.\ are $\downarrow$ (0 = no forgetting); adapt.\ is $\downarrow$. Legal · Qwen3.5-9B has no off-budget max-replay arm (never trained).}
\label{tab:frontier_full}
\resizebox{\linewidth}{!}{
\begin{tabular}{llccccc}
\toprule
Setting & Method & Replay share & Adapt.\ $\downarrow$ & Val.\ forget.\ $\downarrow$ & Task forget.\ $\downarrow$ & Tokens (B) \\
\midrule
Legal $\cdot$ Nemotron-12B
& Base                    & --    & 1.492 & 0.000 & 0.000 & 0.00 \\
& No-replay              & 0\%   & 0.972 & 0.553 & 0.122 & 12.41 \\
& Fixed replay (10\%)     & 10\%  & 0.967 & 0.125 & 0.067 & 12.41 \\
& Fixed replay (20\%)     & 20\%  & 0.970 & 0.082 & 0.055 & 12.41 \\
& Fixed replay (50\%)     & 50\%  & 0.992 & 0.024 & 0.036 & 12.41 \\
& Max-replay (off-budget) & 50\%  & 0.976 & 0.057 & 0.046 & 24.31 \\
& Model merging (0.2)     & --    & 1.225 & 0.038 & 0.000 & 12.41 \\
& Model merging (0.4)     & --    & 1.100 & 0.135 & 0.005 & 12.41 \\
& Model merging (0.6)     & --    & 1.013 & 0.256 & 0.030 & 12.41 \\
& Model merging (0.8)     & --    & 0.970 & 0.390 & 0.064 & 12.41 \\
& \textbf{RoD}           & 21\%  & \textbf{0.967} & \textbf{0.071} & \textbf{0.049} & 12.41 \\
\midrule
German $\cdot$ Nemotron-12B
& Base                    & --    & 2.478 & 0.000 & 0.000 & 0.00 \\
& No-replay              & 0\%   & 1.793 & 0.407 & 0.096 & 12.50 \\
& Fixed replay (17\%)     & 17\%  & 1.810 & 0.122 & 0.067 & 12.50 \\
& Fixed replay (32\%)     & 32\%  & 1.813 & 0.077 & 0.049 & 12.50 \\
& Fixed replay (65\%)     & 65\%  & 1.872 & 0.051 & 0.029 & 12.50 \\
& Max-replay (off-budget) & 65\%  & 1.808 & 0.051 & 0.039 & 24.80 \\
& Model merging (0.2)     & --    & 2.261 & 0.027 & 0.000 & 12.50 \\
& Model merging (0.4)     & --    & 2.076 & 0.095 & 0.005 & 12.50 \\
& Model merging (0.6)     & --    & 1.912 & 0.186 & 0.021 & 12.50 \\
& Model merging (0.8)     & --    & 1.815 & 0.289 & 0.052 & 12.50 \\
& \textbf{RoD}           & 24\%  & \textbf{1.788} & \textbf{0.073} & \textbf{0.050} & 12.50 \\
\midrule
German $\cdot$ Qwen3.5-9B
& Base                    & --    & 2.164 & 0.000 & 0.000 & 0.00 \\
& No-replay              & 0\%   & 1.793 & 0.420 & 0.108 & 12.42 \\
& Fixed replay (17\%)     & 17\%  & 1.796 & 0.035 & 0.052 & 12.42 \\
& Fixed replay (32\%)     & 32\%  & 1.807 & 0.019 & 0.039 & 12.42 \\
& Fixed replay (65\%)     & 65\%  & 1.867 & 0.007 & 0.029 & 12.42 \\
& Max-replay (off-budget) & 65\%  & 1.813 & 0.005 & 0.030 & 24.62 \\
& Model merging (0.2)     & --    & 2.044 & 0.021 & 0.001 & 12.42 \\
& Model merging (0.4)     & --    & 1.938 & 0.091 & 0.006 & 12.42 \\
& Model merging (0.6)     & --    & 1.853 & 0.188 & 0.020 & 12.42 \\
& Model merging (0.8)     & --    & 1.809 & 0.292 & 0.051 & 12.42 \\
& \textbf{RoD}           & 16\%  & \textbf{1.790} & \textbf{0.028} & \textbf{0.061} & 12.42 \\
\midrule
German $\cdot$ Qwen3.5-4B
& Base                    & --    & 2.322 & 0.000 & 0.000 & 0.00 \\
& No-replay              & 0\%   & 1.849 & 0.521 & 0.197 & 12.00 \\
& Fixed replay (17\%)     & 17\%  & 1.863 & 0.036 & 0.090 & 12.00 \\
& Fixed replay (32\%)     & 32\%  & 1.878 & 0.020 & 0.079 & 12.00 \\
& Fixed replay (65\%)     & 65\%  & 1.919 & 0.006 & 0.067 & 12.00 \\
& Max-replay (off-budget) & 65\%  & 1.865 & 0.003 & 0.056 & 27.21 \\
& Model merging (0.2)     & --    & 2.188 & 0.035 & 0.003 & 12.00 \\
& Model merging (0.4)     & --    & 2.060 & 0.140 & 0.019 & 12.00 \\
& Model merging (0.6)     & --    & 1.944 & 0.274 & 0.062 & 12.00 \\
& Model merging (0.8)     & --    & 1.876 & 0.374 & 0.122 & 12.00 \\
& \textbf{RoD}           & 18\%  & \textbf{1.859} & \textbf{0.028} & \textbf{0.107} & 12.00 \\
\midrule
Legal $\cdot$ Qwen3.5-9B
& Base                    & --    & 1.398 & 0.000 & 0.000 & 0.00 \\
& No-replay              & 0\%   & 0.986 & 0.571 & 0.147 & 11.84 \\
& Fixed replay (10\%)     & 10\%  & 0.988 & 0.075 & 0.070 & 11.84 \\
& Fixed replay (20\%)     & 20\%  & 0.992 & 0.044 & 0.055 & 11.84 \\
& Fixed replay (50\%)     & 50\%  & 1.015 & 0.016 & 0.037 & 11.84 \\
& Model merging (0.2)     & --    & 1.219 & 0.041 & 0.000 & 11.84 \\
& Model merging (0.4)     & --    & 1.107 & 0.139 & 0.003 & 11.84 \\
& Model merging (0.6)     & --    & 1.026 & 0.274 & 0.025 & 11.84 \\
& Model merging (0.8)     & --    & 0.986 & 0.412 & 0.076 & 11.84 \\
& \textbf{RoD}           & 15\%  & \textbf{0.988} & \textbf{0.031} & \textbf{0.059} & 11.84 \\
\bottomrule
\end{tabular}
}
\end{table}

\paragraph{Full frontier results.}
\Cref{tab:frontier_full} reports the complete numerical results underlying the adaptation--forgetting frontiers in \cref{fig:frontier}. We report adaptation loss, validation-loss forgetting, and task-based forgetting for all fixed-replay, model-merging, and RoD runs, together with their realized replay shares and training budgets. Except for the explicitly marked max-replay runs, all training-based methods within each setting are trained-token-matched.

\begin{figure*}[t]
    \centering
    \includegraphics[width=\textwidth]{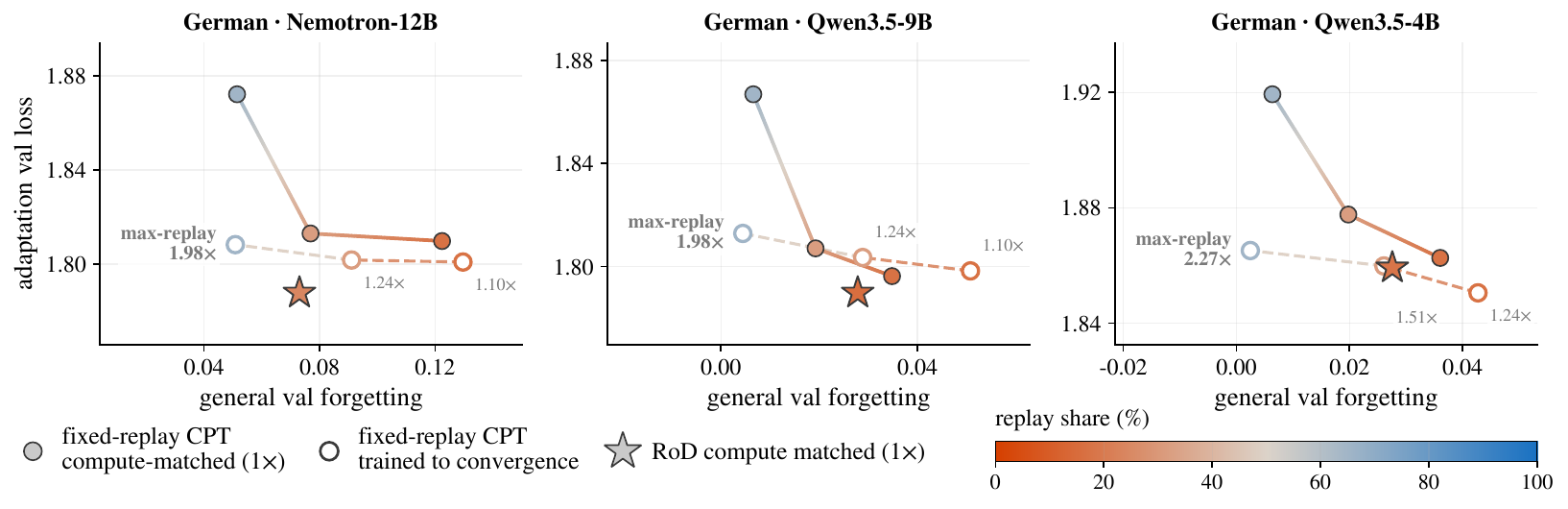}
    \caption{
    \textbf{Fixed-replay CPT trained to convergence.}
    Adaptation--forgetting trade-offs for German adaptation when fixed-replay CPT is trained beyond the matched trained-token ($1\times$) budget until convergence.
    Filled circles show the trained-token-matched fixed-replay runs used in the main comparison, while open circles show the corresponding runs trained to convergence; annotations indicate their training budget relative to the trained-token-matched runs.
    RoD (star) is shown at the $1\times$ compute budget.
    Extending fixed-replay training improves adaptation but does not recover a consistently better adaptation--forgetting trade-off than RoD, while requiring up to $2.27\times$ the training budget.
    }
    \label{fig:frontier_budget}
\end{figure*}

\paragraph{Training fixed-replay baselines to convergence.}
Beyond the trained-token-matched comparison in the main paper, we additionally evaluate fixed-replay CPT, allowing each run to train until convergence.
As shown in \Cref{fig:frontier_budget}, additional training primarily improves adaptation, shifting the fixed-replay operating points downward.
However, these gains require substantially larger training budgets, reaching up to $1.98\times$ for Nemotron-12B and $2.27\times$ for Qwen3.5-4B.
Even with these additional training budgets, RoD at $1\times$ compute remains on or improves upon the resulting adaptation--forgetting frontier.
Thus, the favorable trade-off of RoD is not an artifact of restricting fixed replay to the trained-token-matched budget.

\begin{table}[t]
    \centering
    \small
    \caption{
    \textbf{RoD performance beyond the compute-matched training budget.}
    Continuing RoD beyond the $1\times$ budget does not improve either adaptation or retention, indicating that RoD has already converged at the budget used in our main comparison.
    }
    \label{tab:rod_convergence}
    \begin{tabular}{lccc}
        \toprule
        Setting & Budget & Adapt.\ $\downarrow$ & Val.\ forget.\ $\downarrow$ \\
        \midrule
        German $\cdot$ Nemotron-12B
            & $1.00\times$ & 1.7878 & 0.0731 \\
            & $1.19\times$ & 1.7882 & 0.0734 \\
        \midrule
        Legal $\cdot$ Nemotron-12B
            & $1.00\times$ & 0.9670 & 0.0714 \\
            & $1.20\times$ & 0.9705 & 0.0725 \\
        \bottomrule
    \end{tabular}
\end{table}

We additionally continue RoD for approximately $20\%$ beyond the main training budget for German and Legal adaptation with Nemotron-12B.
As shown in \Cref{tab:rod_convergence}, neither adaptation nor retention improves with additional training, indicating that RoD has already converged at the $1\times$ budget.
Together, these results show that additional training can narrow the adaptation gap for fixed replay, but does not yield a better adaptation--forgetting frontier than RoD despite requiring substantially more compute.

\subsection{Proxy Replay and Task-Based Forgetting}
\label{app:qwen_proxy_replay}

For Nemotron, RoD draws replay examples from the model's own pretraining corpus, such that the pretrained model provides a natural reference for measuring degradation on these data. For Qwen3.5, the original pretraining data are unavailable, and we instead use the Nemotron pretraining corpus as proxy replay. As discussed in \Cref{sec:adaptation_forgetting_frontier}, RoD continues to perform well in terms of validation-loss forgetting in this setting, but its advantage is less pronounced under task-based evaluation. We examine this discrepancy in more detail below.

\paragraph{Qwen is less well fit to the proxy replay distribution.}
A central difference between native and proxy replay is the pretrained model's fit to the replay distribution.
Table~\ref{tab:proxy_replay_base_loss} reports the base-model loss on the held-out replay data, both overall and across the four broad replay categories.
The Nemotron base model achieves lower loss on these data than either Qwen3.5 model.
Its overall replay loss ranges from 0.81 to 0.90 across our Legal and German setups, compared with 1.02--1.03 for Qwen3.5-9B and 1.08 for Qwen3.5-4B.
This difference is consistent across world knowledge, reasoning and mathematics, multilingual data, and code.

This distinction affects how the replay signal should be interpreted.
RoD scores replay candidates relative to the pretrained model,
$\rho_R(x;\theta_t)=\ell_{\theta_t}(x)-\ell_{\theta_0}(x)$.
For native replay, the reference loss is measured on the model's own pretraining distribution and therefore directly captures degradation relative to its pretrained state on that distribution.
For proxy replay, the same base-relative comparison remains a meaningful measure of degradation on the available replay data, but the data are less closely aligned with the distribution on which the model was originally trained.
Consequently, validation-loss forgetting remains informative about retention on the proxy distribution, while its correspondence to changes in the model's original pretraining capabilities may be weaker.
This motivates our complementary task-based evaluation, which directly measures whether the same trends extend to downstream capabilities.

\begin{table}[t]
    \centering
    \small
    \caption{
    \textbf{Base-model loss on the replay distribution.}
    We report held-out language-modeling loss on the Nemotron replay corpus overall and across its four broad source categories. Nemotron is evaluated on replay data from its own pretraining distribution, whereas the same corpus serves as proxy replay for Qwen3.5. The higher Qwen losses indicate that the Qwen base models are less converged on the proxy replay distribution.
    }
    \label{tab:proxy_replay_base_loss}
    \begin{tabular}{lccccc}
        \toprule
        Setting
        & World knowl.
        & Reason./math
        & Multiling.
        & Code
        & Overall \\
        \midrule
        Legal $\cdot$ Nemotron-12B  & 1.30 & 0.49 & 1.07 & 0.40 & 0.81 \\
        German $\cdot$ Nemotron-12B & 1.36 & 0.49 & 1.08 & 0.43 & 0.90 \\
        \midrule
        Legal $\cdot$ Qwen3.5-9B    & 1.47 & 0.60 & 1.51 & 0.53 & 1.03 \\
        German $\cdot$ Qwen3.5-9B   & 1.46 & 0.60 & 1.51 & 0.53 & 1.02 \\
        German $\cdot$ Qwen3.5-4B   & 1.55 & 0.64 & 1.61 & 0.56 & 1.08 \\
        \bottomrule
    \end{tabular}
\end{table}

\begin{figure}[t]
    \centering
    \includegraphics[width=\linewidth]{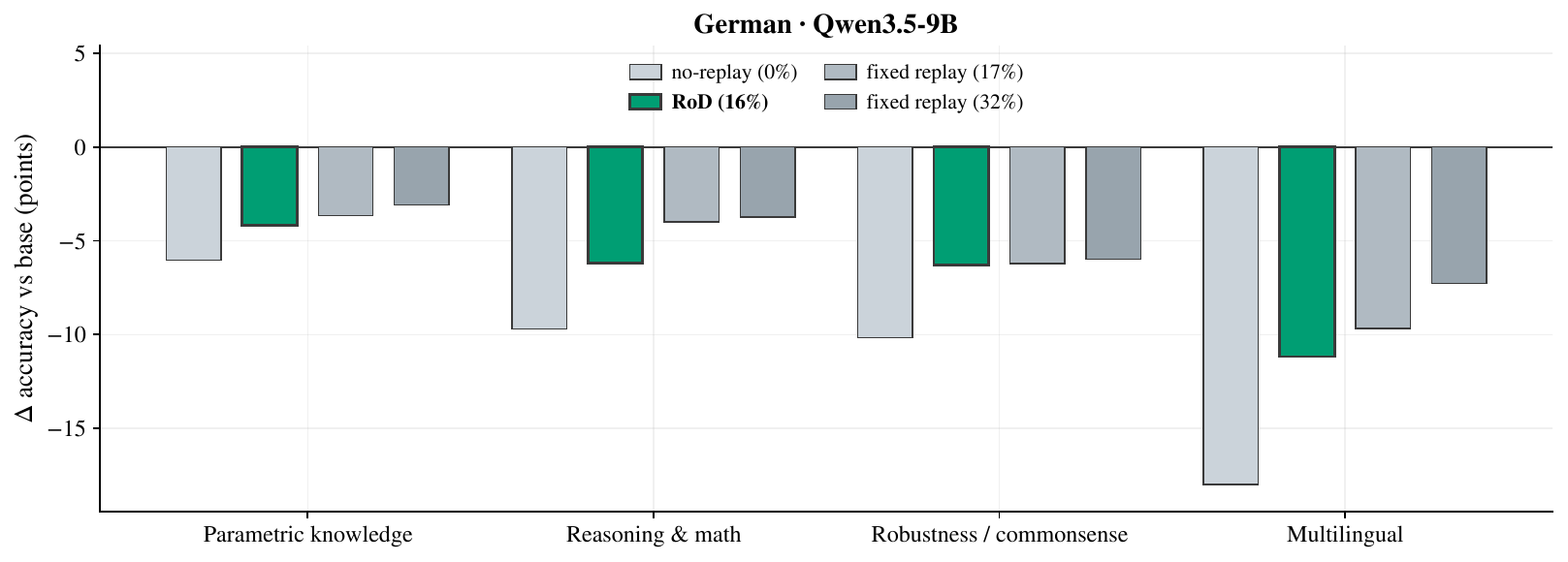}
    \caption{
    \textbf{Task-based forgetting with proxy replay.}
    Change in capability accuracy relative to the pretrained Qwen3.5-9B base model after German adaptation; values closer to zero indicate stronger retention. RoD substantially reduces forgetting relative to no-replay CPT across all capability groups. Its remaining gap to fixed replay is concentrated primarily in reasoning and mathematics and, to a lesser extent, multilingual evaluations, many of which contain translated reasoning and mathematical tasks.
    }
    \label{fig:qwen_task_forgetting}
\end{figure}

\paragraph{The discrepancy is concentrated in reasoning-intensive capabilities.}
We next decompose task-based forgetting by capability to identify where proxy replay diverges from the validation-loss signal. Figure~\ref{fig:qwen_task_forgetting} shows the change in accuracy relative to the pretrained Qwen3.5-9B model for German adaptation. No-replay CPT degrades all four capability groups, with the largest drops in multilingual and robustness/commonsense evaluations. Both RoD and fixed replay substantially mitigate these losses, confirming that proxy replay still provides a useful retention signal.

The remaining difference between RoD and fixed replay is not uniform across capabilities. Parametric knowledge is retained comparatively well by RoD, closely matching the fixed-replay baselines. The clearest gap instead occurs for reasoning and mathematics: RoD reduces the no-replay accuracy drop from approximately 10 points to 6 points, but retains less performance than either fixed-replay baseline. A similar pattern appears in the multilingual group, where RoD improves substantially over no-replay but still lags behind fixed replay. This connection is consistent with the composition of our multilingual evaluations, which contain a substantial fraction of translated mathematical and reasoning tasks. Thus, part of the apparent multilingual forgetting reflects the same reasoning-intensive capabilities for which proxy replay provides weaker protection.

Together, these results clarify the discrepancy between the validation loss and the task-based forgetting observed for Qwen3.5. Proxy replay remains effective; RoD substantially reduces task-based forgetting relative to no-replay CPT, but the replay distribution is less closely matched to the knowledge acquired during Qwen's original pretraining. As a result, base-relative loss changes on the proxy corpus provide a noisier signal of capability retention than under native replay. This limitation is most visible for reasoning-intensive capabilities, where strong validation-loss retention does not translate as directly into task-level retention. We therefore view the Qwen experiments as a more challenging proxy-replay setting rather than a direct test of RoD with access to the model's original pretraining distribution.

\subsection{Demand-Driven Replay Across Model--Domain Settings}
\label{app:demand_driven_extended}

\subsubsection{RHO Loss Dynamics}
\label{app:rho_losses}

\begin{figure*}[t]
    \centering
    \includegraphics[width=\textwidth]{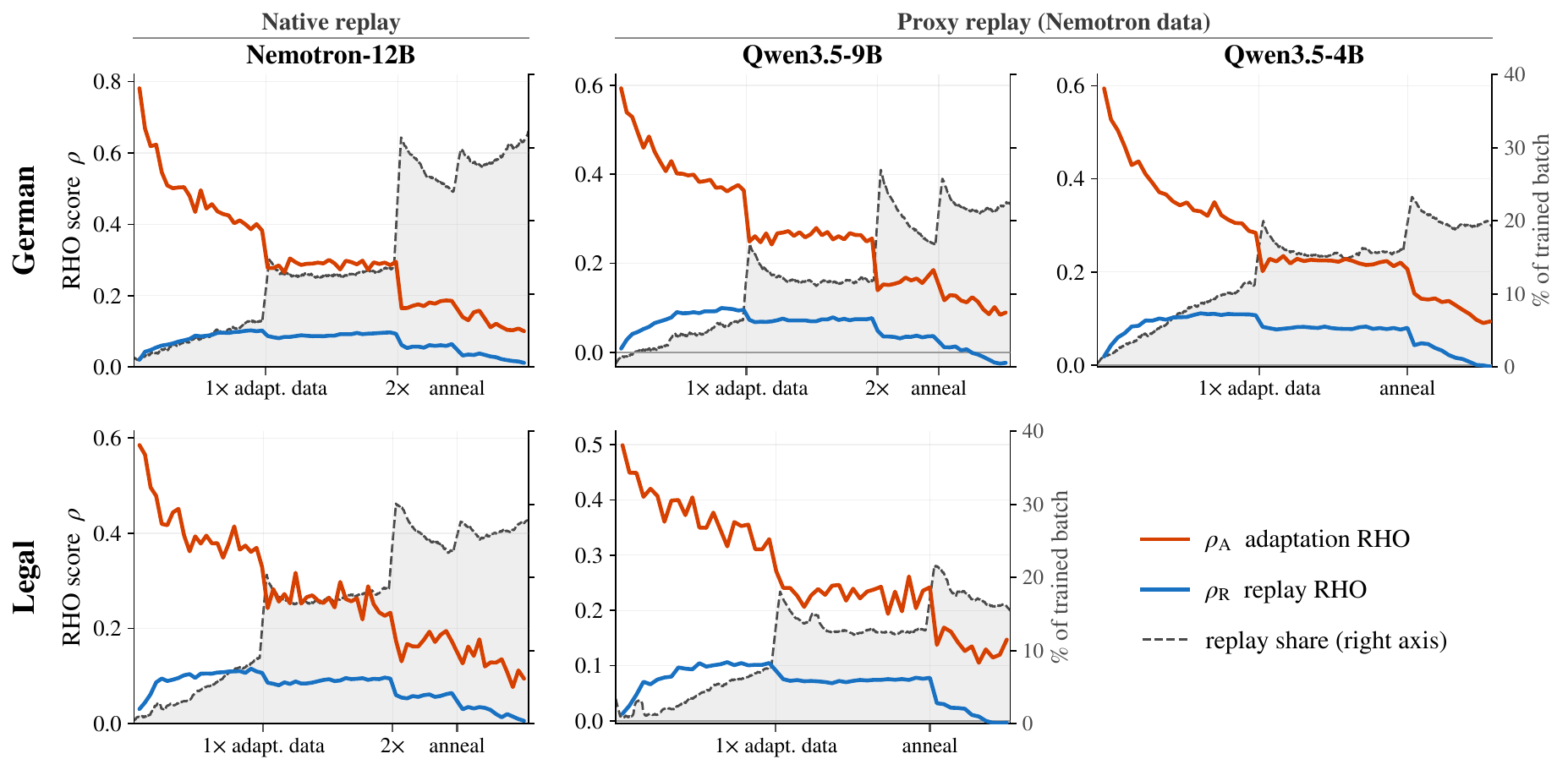}
    \caption{
    \textbf{RHO-score dynamics across model--domain settings.}
    Mean adaptation RHO score $\rho_A$ and replay RHO score $\rho_R$ over training, together with the resulting replay share.
    Across settings, adaptation scores decrease as the model learns the adaptation distribution, while replay scores initially increase as forgetting accumulates.
    The resulting competition between both signals dynamically adjusts the amount of replay throughout training.
    Vertical changes in replay share coincide with transitions between data passes and the learning-rate annealing phase.
    }
    \label{fig:rho_loss_dynamics}
\end{figure*}

Figure~\ref{fig:rho_loss_dynamics} extends the RHO-score dynamics shown in the main paper to the remaining model--domain settings.
Across all settings, we observe the same qualitative feedback loop underlying RoD.
At the beginning of training, adaptation examples have high RHO scores because substantial learning potential remains, whereas replay scores start near zero, as the current model is still close to the pretrained model.
As training progresses, adaptation scores decrease as the model learns the adaptation distribution, while replay scores increase as the model begins to forget.
Replay therefore becomes increasingly competitive in the joint selection, causing the replay share to grow over training.

The abrupt changes in replay share coincide with transitions between passes over the adaptation data and the onset of learning-rate annealing.
At these points, previously seen adaptation examples re-enter the candidate stream, temporarily changing their RHO-score distribution and thereby the balance between adaptation and replay.
Despite these common dynamics, the resulting replay curricula differ across models and domains.
For example, the replay share rises more sharply for Nemotron than for Qwen across several phases, and its trajectory differs between the German and Legal adaptations.
Thus, RoD consistently exhibits the intended feedback mechanism while adapting the resulting amount of replay to the forgetting dynamics of each model--domain setting.

\subsubsection{Forgetting Profiles}
\label{app:forgetting_profiles}

\begin{figure*}[t]
    \centering
    \includegraphics[width=\textwidth]{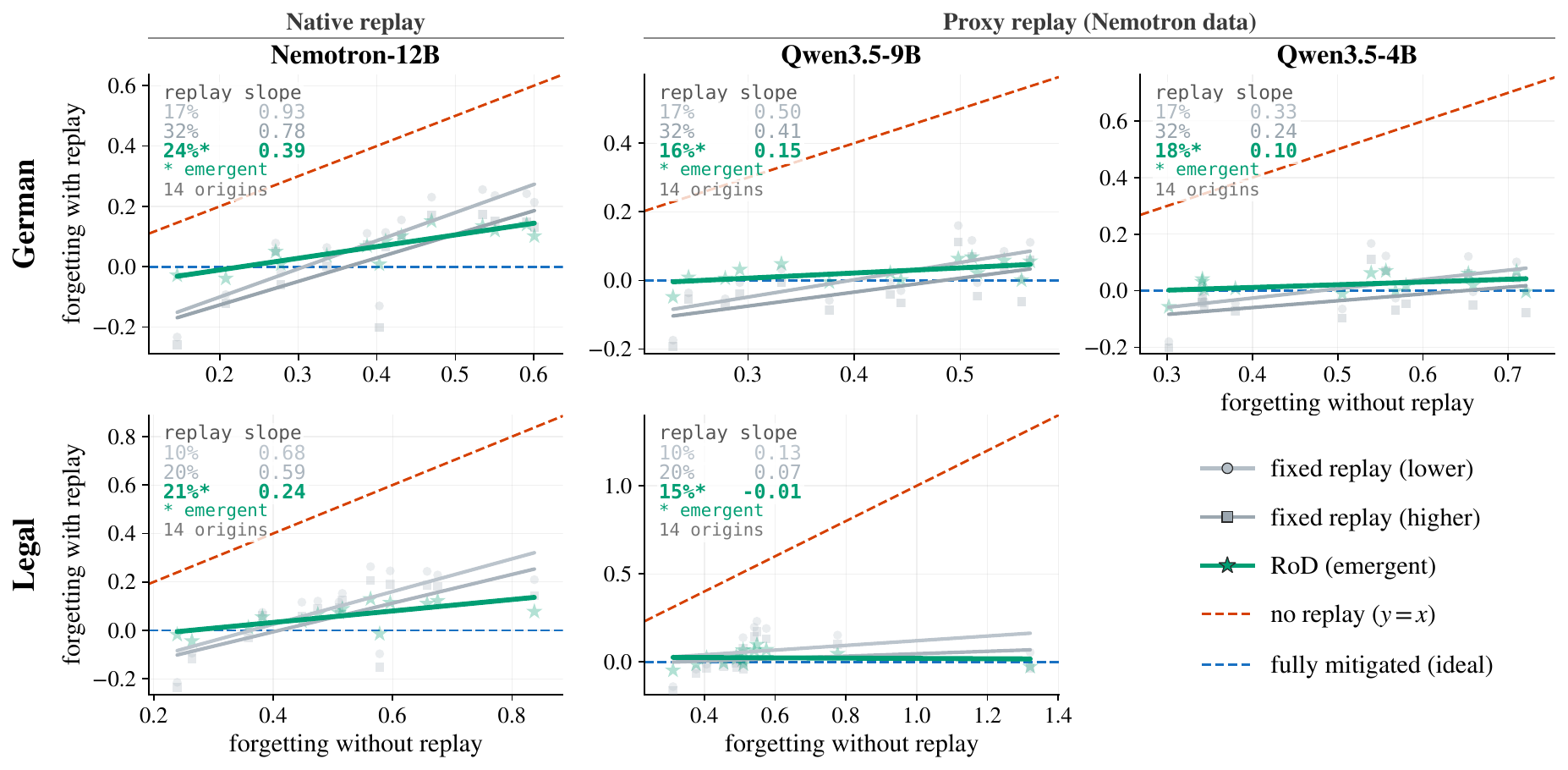}
    \caption{
    \textbf{Source-level forgetting profiles across model--domain settings.}
    Each point represents one replay-data source. The $x$-axis measures source-level forgetting under no-replay CPT, capturing how vulnerable a source is to forgetting, while the $y$-axis measures the forgetting that remains after applying replay.
    Lines show linear fits for fixed-replay CPT and RoD; the reported slopes summarize how strongly remaining forgetting depends on the source's original vulnerability.
    A slope near one indicates approximately uniform reduction of the no-replay forgetting profile, whereas a flatter slope indicates stronger relative protection of sources that would otherwise forget most.
    Across settings, RoD consistently produces the flattest forgetting profile.
    }
    \label{fig:forgetting_profiles_extended}
\end{figure*}

To examine whether the source-level forgetting patterns observed in the main paper generalize across settings, Figure~\ref{fig:forgetting_profiles_extended} reports the corresponding profiles for the remaining model–domain combinations.
For each replay source, we use forgetting under no-replay CPT as a measure of its vulnerability to continual adaptation ($x$-axis) and compare it against the forgetting remaining after replay ($y$-axis).
The slope of the resulting forgetting profile captures how strongly this initial vulnerability persists after replay.
A slope close to one indicates that replay approximately shifts the no-replay profile downward while preserving the relative differences between sources; a flatter slope indicates that sources prone to forgetting receive disproportionately stronger protection.

We observe the same qualitative pattern across models and adaptation domains.
Fixed-replay CPT reduces overall forgetting, but largely preserves the underlying forgetting profile: sources that forget most without replay generally remain those with the largest residual forgetting.
Increasing the fixed replay share pushes this profile further downward and moderately reduces its slope.
RoD changes the profile more substantially.
Across all settings, it yields the flattest fitted slope, preferentially reducing forgetting for vulnerable sources while spending less replay capacity on sources that already remain stable.
For example, for German adaptation the slope decreases from $0.93/0.78$ under fixed replay to $0.39$ with RoD for Nemotron-12B, from $0.50/0.41$ to $0.15$ for Qwen3.5-9B, and from $0.33/0.24$ to $0.10$ for Qwen3.5-4B.
The same behavior appears for Legal adaptation, where RoD reduces the slope to $0.06$ for Nemotron-12B.

The Legal--Qwen3.5-9B setting contains one influential multilingual knowledge category that substantially flattens all fitted profiles.
We retain this knowledge category in \cref{fig:forgetting_profiles_extended}, but verify that the result is not driven by this outlier.
Excluding it increases the slopes of the two fixed-replay profiles from $0.13$ and $0.07$ to $0.74$ and $0.65$, respectively, while the RoD slope increases from $-0.01$ to only $0.23$.
Thus, the same pattern becomes clearer after excluding the outlier: fixed replay primarily reduces the magnitude of forgetting while retaining much of the source-level vulnerability profile, whereas RoD more strongly redistributes protection toward the sources that are most susceptible to forgetting.

Overall, these results show that the demand-driven behavior observed for German adaptation with Nemotron-12B generalizes across model families and domains.
RoD does not merely determine how much replay to use; its joint selection also changes which parts of the pretraining distribution are protected, concentrating replay where forgetting is most pronounced.

\subsubsection{Replay Allocation}
\label{app:oversampling}

\begin{figure*}[t]
    \centering
    \includegraphics[width=\textwidth]{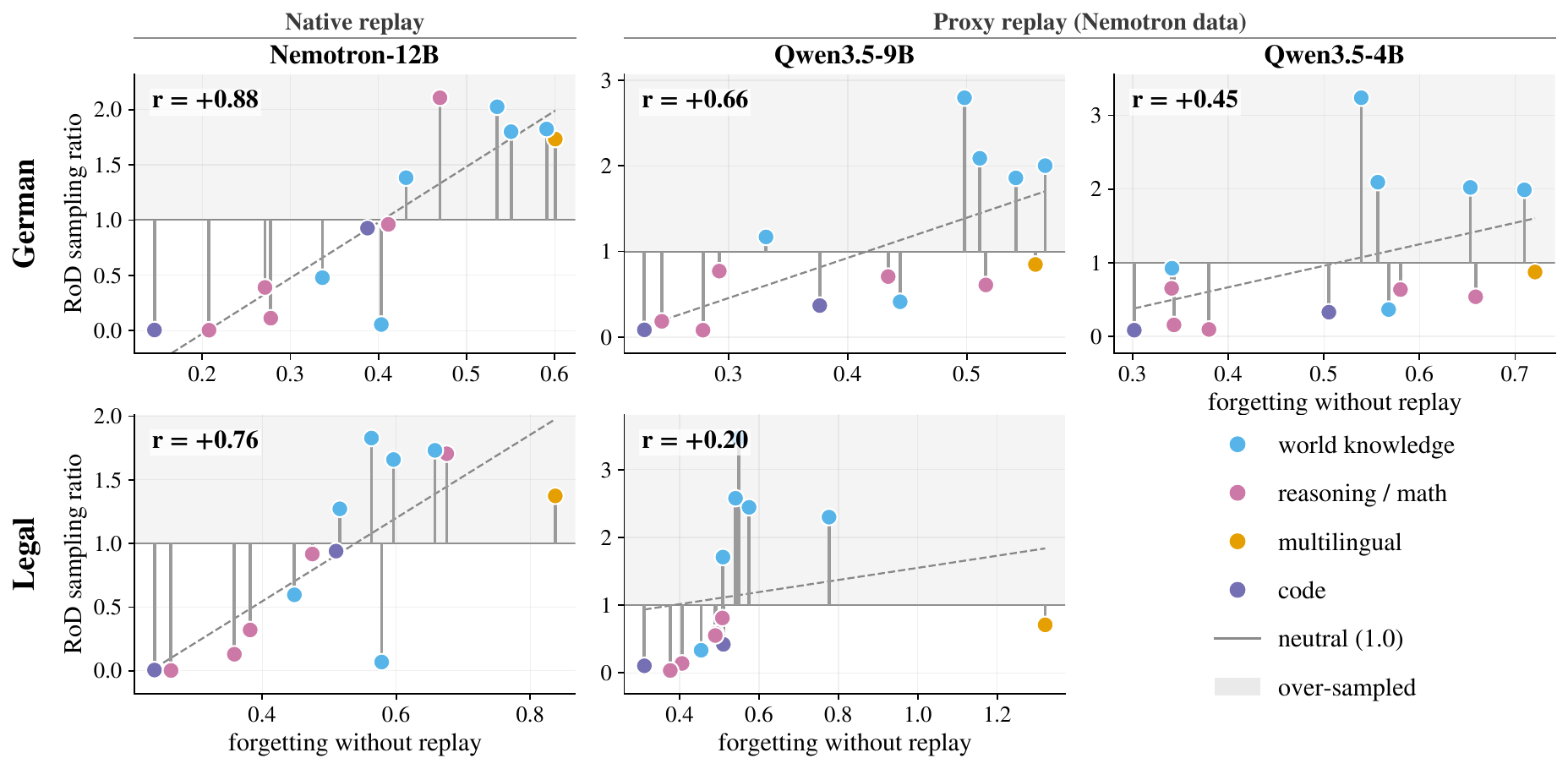}
    \caption{
    \textbf{Replay allocation follows source-level forgetting across model--domain settings.}
    Each point represents one replay-data source. The $x$-axis measures source vulnerability as forgetting under no-replay CPT, while the $y$-axis shows its sampling ratio under RoD relative to its prevalence in the replay candidate distribution; values above $1$ indicate oversampling and values below $1$ undersampling.
    Colors denote broad pretraining-data categories.
    Dashed lines show linear fits and $r$ denotes the Pearson correlation between vulnerability and RoD sampling ratio.
    Across settings, RoD preferentially allocates replay to sources that are more susceptible to forgetting.
    }
    \label{fig:oversampling_extended}
\end{figure*}

The flatter forgetting profiles in \cref{app:forgetting_profiles} suggest that RoD allocates replay preferentially toward sources that are most vulnerable to forgetting.
Figure~\ref{fig:oversampling_extended} directly tests this mechanism across the remaining model--domain settings.
For each replay source, we compare its vulnerability (measured as forgetting under no-replay CPT) against its sampling ratio under RoD.
A ratio of $1$ corresponds to sampling proportional to the source's prevalence in the replay candidate distribution, while values above or below $1$ indicate increased or decreased sampling frequency, respectively.

Across settings, replay allocation is positively associated with source vulnerability.
The relationship is strongest for German adaptation with Nemotron-12B ($r=0.88$), but remains positive for Qwen3.5-9B ($r=0.66$), Qwen3.5-4B ($r=0.45$), and Legal adaptation with Nemotron-12B ($r=0.76$).
Sources that remain comparatively stable are often sampled below their prevalence in the replay distribution, whereas highly vulnerable sources can be replayed at more than twice their baseline rate.
Thus, the source-level protection observed in \cref{app:forgetting_profiles} arises from an adaptive redistribution of replay rather than from uniformly increasing replay across the pretraining distribution.

As in the forgetting-profile analysis, the Legal--Qwen3.5-9B setting contains an influential multilingual source.
Including this source yields a weaker correlation of $r=0.20$; excluding it increases the correlation to $r=0.69$ across the remaining 13 sources.

Together, these results show that the demand-driven allocation observed in the main-paper setting generalizes across models and domains: RoD uses the emerging forgetting signal not only to determine \emph{how much} to replay, but also \emph{what} to replay, concentrating its replay budget on the parts of the pretraining distribution that currently require it most.

% ============================================================
\section{Limitations and Conceptual Considerations}
\label{app:limitations}
% ============================================================

\subsection{Computational Requirements}
\label{app:computational_requirements}

RoD introduces additional computation relative to standard CPT through both online candidate scoring and offline reference construction.
During training, each optimization step first scores a candidate pool of $mk$ examples using a forward-only pass, before applying the standard forward--backward update to the selected batch of $k$ examples.
Candidate scoring requires no gradients or optimizer updates, but its cost grows linearly with the candidate multiplier $m$.
In our experiments, we use $m=2$, such that $2k$ candidates are evaluated for every $k$ examples used for training.
Our ablations further show little benefit from increasing the candidate pool beyond $m=2$, limiting the additional online computation required in our experiments.

RoD additionally requires reference losses for both candidate streams.
For replay, the reference is the pretrained model itself; for adaptation, our implementation constructs a specialized reference by training on the adaptation corpus until convergence.
In our experiments, this specialist also serves as the no-replay CPT baseline and therefore does not require a separate training run within the experimental suite.
In a standalone RoD application, however, constructing such a reference incurs an additional offline cost unless a suitable specialist is already available.
Reference losses are fixed and can be precomputed and cached, such that neither reference model needs to be evaluated during RoD training.
Moreover, the adaptation reference need not match the scale of the model being adapted: our cross-scale ablation shows that a 4B reference recovers a similar adaptation--forgetting trade-off for a 9B target model.

Our implementation is designed primarily to study whether dynamic, loss-based replay allocation is effective and to characterize the resulting data curriculum, rather than to optimize computational efficiency.
Importantly, our cross-scale experiments provide a practical route to decoupling RoD's computational requirements from the target model's scale.
A curriculum constructed by a smaller model can be transferred to a larger target model, avoiding candidate scoring and model-dependent reference-loss computation at the target model's scale.
Together with the smaller adaptation reference above, these results suggest that RoD's model-dependent computation need not scale directly with the model being adapted.
Further reducing the cost of curriculum construction remains an important direction for future work.

\subsection{Relative Weighting of Adaptation and Replay}
\label{app:relative_weighting}

RoD directly compares adaptation and replay scores in the same token-normalized loss space, placing both signals on equal footing during joint selection.
This provides a parameter-free default in which the adaptation--retention trade-off emerges from the model's current learning and forgetting state rather than from a predefined preference.
For applications that require a specific operating point, the framework could be extended with a relative weighting between the two signals to explicitly favor adaptation or retention.
We leave this extension to future work.

\subsection{Requirements on the Replay Distribution}
\label{app:replay_distribution}

RoD can only protect knowledge that is represented in the available replay distribution.
While the method determines which replay examples to prioritize and how much replay to allocate during training, it assumes access to a replay pool that sufficiently covers the parts of the pretraining distribution that should be retained.
If relevant knowledge or capabilities are absent from this pool, their degradation cannot be detected through the corresponding replay examples, and RoD cannot selectively allocate replay toward them.
Thus, RoD removes the need to prescribe the composition and amount of replay within an available replay distribution, but does not remove the requirement for representative replay data.

Our Qwen3.5 experiments illustrate that this requirement does not imply access to the exact original pretraining corpus.
Because Qwen's pretraining data are unavailable, we use the Nemotron corpus as proxy replay and still observe substantial reductions in forgetting.
At the same time, the larger discrepancy between validation loss and task-based forgetting in this setting indicates that mismatched replay data can yield a less faithful retention signal.
We therefore expect RoD to be most effective when the replay distribution provides broad coverage of the knowledge and capabilities that should be retained, while representative proxy data remain a viable alternative when the original pretraining distribution is unavailable.

\subsection{Interpretation and Scope of the Replay Signal}
\label{app:replay_signal}

RoD operationalizes forgetting through increases in language-modeling loss relative to the pretrained model,
$\rho_R(x;\theta_t)=\ell_{\theta_t}(x)-\ell_{\theta_0}(x)$.
This provides a task-agnostic signal that can be evaluated directly on replay examples during CPT, without requiring downstream labels or capability-specific evaluations.
However, sequence-level loss degradation is not equivalent to downstream capability degradation.
RoD can therefore react only to forgetting that manifests as increased loss on the available replay data, and the strength of this correspondence depends on how well those data represent the capabilities of interest.

This distinction is most visible in our proxy-replay experiments.
For Nemotron, replay examples originate from the model's own pretraining distribution, providing a direct reference for degradation from the pretrained state.
For Qwen3.5, loss changes are instead measured using the Nemotron proxy distribution, and we observe a weaker correspondence between validation loss and task-based forgetting, particularly for reasoning-intensive capabilities.
Nevertheless, RoD substantially reduces task-based forgetting relative to no-replay CPT in these settings, suggesting that the loss-based signal remains useful even when this correspondence is imperfect.
More generally, our results support base-relative loss as a practical online signal for demand-driven replay, while its interpretation as capability forgetting should remain tied to the coverage and alignment of the replay distribution.

\end{document}